\documentclass[a4paper,fleqn]{cas-dc}
\usepackage[square, numbers]{natbib}
 
\usepackage{caption}
\usepackage{cas-common}
\usepackage{graphicx}
\usepackage{diagbox}
\usepackage{bbding}
\usepackage{pifont}
\usepackage{wasysym}
\usepackage{tikz}
\usepackage{calc}
\usepackage{multirow}
\usepackage{rotating}
\usepackage{colortbl}
\usepackage{color}
\usepackage{ragged2e}
\usepackage{float}
\usepackage[inkscapelatex=false]{svg}
\usepackage{latexsym}
\usepackage{booktabs}
\usepackage{pifont}
\usepackage{subcaption} 
\usepackage{svg}
\usepackage{amssymb}
\usepackage{cleveref} 
\usepackage{tikz}
\usetikzlibrary{positioning}
\usepackage{algorithm}
\usepackage{xcolor}
\usepackage{adjustbox}
\usepackage{wrapfig}
\usepackage{algpseudocode}
\usepackage{tabularx}
\usepackage{tcolorbox}
\usepackage{caption}
\usepackage{float}
\usepackage{lipsum}
\usepackage{xtab}
\usepackage{makecell}
\usepackage{xcolor}
\definecolor{lowcontrastgreen}{RGB}{160,206,86}
\definecolor{lowcontrastred}{RGB}{255,95,83}
\definecolor{lowcontrastyellow}{RGB}{255,218,93}
\definecolor{lowcontrastblue}{RGB}{181,234,215}
\usepackage[T1]{fontenc}
\usepackage{microtype}
\usepackage{multicol}
\usepackage{multirow}
\usepackage{inconsolata}
\usepackage{svg}
\usepackage{graphicx}
\usepackage{listings}
\usepackage{epsfig,endnotes}
\usepackage[utf8]{inputenc}
\usepackage[T1]{fontenc}
\usepackage{upquote}
\usepackage[edges]{forest}

\begin{document}
\captionsetup[figure]{labelfont={bf},labelformat={default},name={Fig.}, labelsep=period,}
\captionsetup[table]{labelfont={bf},labelformat={default},labelsep=newline,name={Table}, singlelinecheck=false}
\let\WriteBookmarks\relax
\def\floatpagepagefraction{1}
\def\textpagefraction{.001}

\shorttitle{Embodied Intelligence for 3D Understanding: A Survey on 3D Scene Question Answering}    

\shortauthors{Z. Li et~al. }  

\title [mode = title]{Embodied Intelligence for 3D Understanding: A Survey on 3D Scene Question Answering}  



%


\author[1,2]{Zechuan Li}[type=editor, auid=000, bioid=1,
orcid= 0009-0003-4715-703X
]
\ead{lizechuan@hnu.edu.cn}

\author[1]{Hongshan Yu}[type=editor,auid=000,bioid=1, 
orcid = 0000-0003-1973-6766
]
\cormark[1]
\ead{yuhongshancn@hotmail.com}
\cortext[1]{Corresponding author.}

\author[3]{Yihao Ding}[type=editor,auid=000,bioid=1,]
\ead{yihao.ding@sydney.edu.au}

\author[3]{Yan Li}[type=editor,auid=000,bioid=1,]
\ead{yali3816@uni.sydney.edu.au}

\author[4]{Yong He}[type=editor,auid=000,bioid=1,]
\ead{h.yong@hnu.edu.cn}

\author[2]{Naveed Akhtar}[type=editor,auid=000,bioid=1,]
\ead{naveed.akhtar1@unimelb.edu.au}

\affiliation[1]{organization={College of Electrical and Information Engineering,Hunan University},
            city={Changsha},
            postcode={410082}, 
            state={Hunan},
            country={China}}   

\affiliation[2]{organization={School of Computing \& Information Systems ,The University of Melbourne},
            city={Melbourne},
            postcode={VIC 3053}, 
            state={VIC},
            country={Australia}}  
            
\affiliation[3]{organization={School of Computer Science,The University of Sydney},
            city={Sydney},
            postcode={NSW 2006}, 
            state={NSW},
            country={Australia}}

\affiliation[4]{organization={School of Artificial Intelligence ,Anhui University},
            city={Hefei},
            postcode={230601}, 
            state={Anhui},
            country={China}}

\begin{abstract}
3D Scene Question Answering (3D SQA) represents an interdisciplinary task that integrates 3D visual perception and natural language processing, empowering intelligent agents to comprehend and interact with complex 3D environments. Recent advances in large multimodal modelling have driven the creation of diverse datasets and spurred the development of instruction-tuning and zero-shot methods for 3D SQA. However, this rapid progress introduces challenges, particularly in achieving unified analysis and comparison across datasets and baselines. 
In this survey, we provide the first comprehensive and systematic review of 3D SQA. We organize existing work from three perspectives: datasets, methodologies, and evaluation metrics. Beyond basic categorization, we identify shared architectural patterns across methods. Our survey further synthesizes core limitations and discusses how current trends—such as instruction tuning, multimodal alignment, and zero-shot—can shape future developments. Finally, we propose a range of promising research directions covering dataset construction, task generalization, interaction modeling, and unified evaluation protocols. This work aims to serve as a foundation for future research and foster progress toward more generalizable and intelligent 3D SQA systems.
\end{abstract}




\maketitle

\section{Introduction}
Visual Question Answering (VQA)~\cite{zhang2019information,ishmam2024image}  expands the scope of traditional text-based question answering \cite{squad,bohus2009models,misu2014situated} by incorporating visual content, enabling the interpretation of images \cite{antol2015vqa}, charts \cite{masry2022chartqa}, and documents \cite{ding2024mmvqa} to deliver context-aware responses. This capability facilitates a broader range of applications, including medical diagnostics \cite{wu2022medical}, financial analysis \cite{xue2024famma}, and assistance in academic research. 

\textcolor{black}{While these efforts focus primarily on 2D representations, a growing body of research in 3D visual perception has highlighted the importance of understanding spatial structures and geometric relationships in real-world environments~\cite{qi2017pointnet,qi2017pointnet++}. 
Meanwhile, the emergence of embodied intelligence has brought new demands for agents that can perceive, reason, and communicate within their 3D surroundings~\cite{yuan2025empowering}. Applications such as household robotics~\cite{bai2025surgical,liu2025screens}, AR/VR assistants~\cite{yong2025intervention}, and autonomous navigation~\cite{luo2024transformer} require systems that not only interpret static scenes, but also support interactive multimodal understanding—comprehending user queries in context.}

\begin{figure}[t]
    \centering
    \tiny \includegraphics[width=\linewidth]{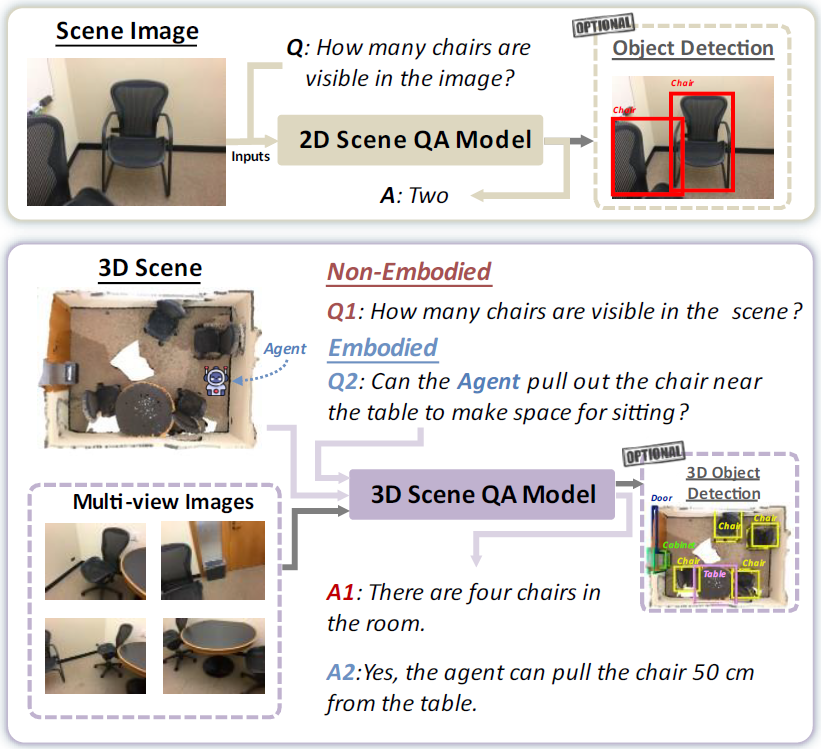}
    \caption{2D Scene VQA and 3D SQA tasks.  3D SQA handles non-embodied as well as embodied tasks involving agent interactions within \textcolor{black}{3D scenes.}} 
    \label{fig:fig1}

\end{figure}

To meet these demands, 3D SQA(Scene Question Answering)~\cite{azuma2022scanqa,ye2021tvcg3dqa}  has emerged as a key task that unifies spatial perception and language understanding. Unlike traditional 3D tasks that focus on object recognition and classification (e.g., 3D object detection~\cite{qi2019deep,li2023ashapeformer} or segmentation~\cite{he2025deep,zou2024improved}), 
3D SQA addresses this by bridging visual perception~\cite{he2016deep,he2017mask}, spatial reasoning~\cite{guo2020deep}, and language understanding in 3D environments \cite{linghu2024multi}. See Figure~\ref{fig:fig1}, 3D SQA integrates multimodal data, e.g., visual inputs and textual queries, to enable embodied systems capable of complex reasoning~\cite{szymanska2024space3d}. By leveraging spatial relationships, object interactions, and hierarchical scene structures within dynamic 3D environments, 3D SQA advances robotics, augmented reality, and autonomous navigation~\cite{huang2023embodied}, pushing the boundaries of multimodal AI and its potential in complex, real-world scenarios.

Early developments in 3D SQA were driven by manually annotated datasets like ScanQA~\cite{azuma2022scanqa} and SQA~\cite{ye2021tvcg3dqa}, which aligned 3D point clouds with textual queries. 
Recently, programmatic generation methods, such as those used in 3DVQA~\cite{9866910} and MSQA~\cite{linghu2024multi}, have enabled the creation of larger datasets with richer question types. The integration of Large Vision-Language Models (LVLMs) has further automated data annotation, leading to the development of more comprehensive datasets like LEO~\cite{huang2023embodied} and  Spartun3D~\cite{zhang2024spartun3d}.

Methodologies have evolved alongside datasets, transitioning from closed-set approaches to LVLM-enabled techniques. Early methods~\cite{azuma2022scanqa,ye2021tvcg3dqa} employed custom architectures combining point cloud encoders, e.g., PointNet++ \cite{qi2017pointnet++}, and text encoders, e.g., BERT \cite{kenton2019bert}, with attention-based fusion modules. However, they were constrained by predefined answer sets. The recent LVLM-based methods employ instruction-tuning~\cite{hong20233d3dllm,huang2023embodied} or zero-shot technique~\cite{yin2024lamm,linghu2024multi} while adapting models like GPT-4~\cite{achiam2023gpt}, which reduces  dependence on task-specific annotations. However, these methods also face  challenges in ensuring dataset quality and addressing evaluation inconsistencies.

To analyse the emerging challenges in 3D SQA and facilitate their systematic handling, this paper provides the first comprehensive survey of this research direction.
We focus on three fundamental aspects of this area, namely; \textit{(i)} the objectives of 3D SQA, \textit{(ii)} datasets needed to support these objectives, and \textit{(iii)} models being developed to achieve these objectives.
We review the evolution of datasets and methodologies, highlighting trends in the literature, such as the shift from manual annotation to LVLM-assisted generation, and the progression from closed-set to zero-shot methods. Additionally, we discuss challenges in multimodal alignment and evaluation standardization, offering insights into the future direction of the field. 


\textcolor{black}{To provide a clear and coherent overview of the field, we organize this survey to follow the developmental trajectory of 3D SQA itself. Since this task evolved from earlier QA and VQA paradigms, its progress has been shaped by the co-evolution of datasets, evaluation benchmarks, and modeling techniques. Accordingly, Section~\ref{sec2} introduces the task and its foundations, followed by Section~\ref{sec3} on dataset construction, Section~\ref{sec4} on evaluation metrics, Section~\ref{sec5} on representative methods, and Section~\ref{sec6} on open challenges. An overview of this organization is provided in Figure~\ref{fig:3d_sqa_structure}.}

\begin{figure*}[htp]
\scriptsize 
\centering
\begin{forest}
    for tree={
        forked edges,
        grow'=0,
        draw,
        rounded corners,
        node options={align=center},
        text width=2.2cm, 
        s sep=5pt, 
        calign=child edge,
        calign child=(n_children()+1)/2, 
    }
    [\textbf{3D SQA }, 
    text width=3cm, 
    fill=gray!30,
    for tree={fill=brown!45}
        [Preliminaries, for tree={fill=orange!30}]
        [Datasets, for tree={fill=blue!40}
            [Dataset Structure, for tree={fill=blue!20}
                [Scene Modalities and Scale, for tree={fill=blue!10}
                    [Synthetic 3D Datasets]
                    [Point Cloud Datasets]
                    [Multi-View Datasets]
                    [Multimodal Datasets]
                ]
                [Query Modalities and Complexity, for tree={fill=blue!10}
                    [Basic Text Queries]
                    [Agent-Centric Text Queries]
                    [Multimodal Agent-Centric Queries]
                    [Instruction-Tuned Queries]
                ]
            ]
            [QA Pair Creation, for tree={fill=blue!20}
                [Methods for QA Pair Generation
                    [Template-Based Generation]
                    [Manual Annotation]
                    [LLM-Assisted Generation]
                ]
                [Question Design in 3D SQA
                [Task Complexity]
                    [Situated vs. Non-Situated]
                    [Temporal Aspec]
                ]
            ]
            [Evaluating LLM-Generated 3D Datasets, for tree={fill=blue!20}]
        ]
        [Evaluation Metrics, for tree={fill=green!40}
            [Traditional metrics]
            [LLM-based metrics]
        ]
        [Methodological Taxonomy, for tree={fill=violet!40}
            [Task-Specific Methods, for tree={fill=violet!20}
                [Point Cloud Methods]
                [Multi-view and 2D-3D Methods]
                [Advances in Text Encoders]
            ]
            [Pretraining-Based Methods, for tree={fill=violet!20}
                [Traditional Pretraining Approaches]
                [Instruction-Tuning Methods]
            ]
            [Zero-Shot Learning Methods, for tree={fill=violet!20}
                [Text-Driven Approaches]
                [Image-Driven Approaches]
                [Multimodal Alignment Approaches]
            ]
            [State-of-the-arts, for tree={fill=violet!20}] 
        ]
        [Challenges and Future Directions, for tree={fill=red!40}
            [Dataset Quality ]
            [Zero-Shot]
            [Unified Evaluation]
            [Dynamic and Open-World, text width=4.2cm] 
            [Interpretable and Explainable, text width=4.2cm]
            [Multimodal and Collaboration, text width=4.2cm]
            [Incorporating Temporal Dynamics, text width=4.2cm]
        ]
    ]
\end{forest}
\caption{Graphical illustration of the hierarchical structure of 3D SQA literature adopted in this work. A systematic categorization is adopted for preliminaries, datasets,  evaluation metrics and methodologies.}
\label{fig:3d_sqa_structure}
\vspace{-3mm}
\end{figure*}

\begin{table}[h]
\footnotesize
\centering
\renewcommand{\arraystretch}{0.7}
\caption{3D SQA task notations.}
\adjustbox{width=0.5\textwidth}{
\begin{tabular}{>{\centering\arraybackslash}m{1.7cm}>{\centering\arraybackslash}m{5.5cm}}
\toprule
\textbf{Notation} & \textbf{Definition} \\ \midrule 
\( S \) & A 3D scene representation. \\ \cmidrule{2-2}
\( S^{(p)} \)  & Point cloud representation of the scene: \( S^{(p)} = \{ \mathbf{x}_i \}_{i=1}^{N},\ \mathbf{x}_i \in \mathbb{R}^3 \). \\ \cmidrule{2-2}
\( S^{(m)} \) & Multi-view RGB image representation: \( S^{(m)} = \{I_1, I_2, \dots, I_K\} \). \\ \midrule
\( Q \) & Multimodal query input. \\ \cmidrule{2-2}
\( Q^{(t)} \) & Textual query: \( Q^{(t)} = (w_1, w_2, \dots, w_L) \). \\ \cmidrule{2-2}
\( Q^{(e)} \) & Egocentric image(s) as part of the query. \\ \cmidrule{2-2}
\( Q^{(o)} \) & Object-level point cloud fragments: \( Q^{(o)} = \{ \mathbf{x}_j \}_{j=1}^{M},\ \mathbf{x}_j \in \mathbb{R}^3 \). \\ \midrule
\( T \) & Textual answer: \( T = (t_1, t_2, \dots, t_R) \). \\ \midrule
\( B^{(3D)} \) & Set of 3D bounding boxes: \( B^{(3D)} = \{ b_1, b_2, \dots, b_M \} \). \\ \midrule
\( \mathcal{F}_{3D} \) & 3D SQA function mapping input to output: \( \mathcal{F}_{3D} : \left( S, Q \right) \mapsto \left( T, B^{(3D)} \right) \). \\
\bottomrule
\end{tabular}
}
\label{tab:notation}
\end{table}



\section{Preliminaries\label{sec2}}

\textcolor{black}{To define the 3D SQA task, we briefly review the progression of question answering paradigms across increasing levels of modality. From text-based QA to visual QA and finally to 3D scene QA, this evolution reflects a growing need for multimodal and spatially grounded understanding.
3D SQA extends prior QA settings by requiring agents to reason over spatially structured scenes—represented by point clouds or multi-view images—and respond to multimodal queries that may include text, egocentric views, or object-level inputs. This section formalizes the task setting, laying the foundation for our discussion of datasets, methods, and evaluation.}

\textbf{Text-based QA} involves answering a textual query \( Q^{(t)} \) based on a given textual passage \( P \), yielding a textual response \( T \):
\begin{equation}
\mathcal{F}_{{QA}} : \left( P, Q^{{(t)}} \right) \mapsto T
\end{equation}

\textbf{Visual QA} incorporates a 2D image \( I \) as context, alongside a textual question. In some cases, the output includes not only an answer \( T \) but also a set of 2D bounding boxes \( B^{{(2D)}} \) for visual grounding:
\begin{equation}
\mathcal{F}_{{VQA}} : \left( I, Q^{{(t)}} \right) \mapsto \left( T, B^{{(2D)}} \right)
\end{equation}

\textbf{3D SQA} further extends this formulation by introducing a 3D scene \( S \), which may consist of point clouds \( S^{{(p)}} \), multi-view RGB images \( S^{{(m)}} \), or both. Moreover, the query \( Q \) itself may be multimodal—combining natural language \( Q^{{(t)}} \), egocentric observations \( Q^{{(e)}} \), or object-level 3D inputs \( Q^{{(o)}} \). The system predicts a textual answer \( T \), and optionally, a set of 3D bounding boxes \( B^{{(3D)}} \) for spatial grounding:
\begin{equation}
\mathcal{F}_{{3D}} : \left( S, Q \right) \mapsto \left( T, B^{{(3D)}} \right)
\end{equation}

This evolution from text QA to embodied 3D SQA reflects a broader shift toward situated and interactive intelligence. As tasks move from textual to visual to spatial contexts, both the input modalities and the expected outputs become increasingly multimodal and grounded.

In 3D SQA, the input consists of a 3D scene \( S \) and a multimodal query \( Q \). The scene \( S \) may include point cloud data and multi-view images. Specifically, the point cloud is denoted as:
\begin{equation}
S^{{(p)}} = \left\{ \mathbf{x}_i \right\}_{i=1}^N, \quad \mathbf{x}_i \in \mathbb{R}^3,
\end{equation}
where each \( \mathbf{x}_i \) represents a 3D coordinate. The multi-view RGB image input is represented as:
\begin{equation}
S^{{(m)}} = \{ I_1, I_2, \dots, I_K \}.
\end{equation}

The query \( Q \) may also be composed of multiple modalities. A textual question is represented as a sequence of tokens:
\begin{equation}
Q^{{(t)}} = (w_1, w_2, \dots, w_L).
\end{equation}
In addition, the query may include egocentric visual observations \( Q^{{(e)}} \), typically corresponding to first-person views of the scene, as well as object-level point cloud fragments:
\begin{equation}
Q^{{(o)}} = \left\{ \mathbf{x}_j \right\}_{j=1}^M.
\end{equation}

The output of the system includes a natural language answer:
\begin{equation}
T = (t_1, t_2, \dots, t_R),
\end{equation}
and, optionally, a set of 3D bounding boxes indicating spatial grounding:
\begin{equation}
B^{{(3D)}} = \{ b_1, b_2, \dots, b_M \}.
\end{equation}

Key notations are summarized in Table~\ref{tab:notation}.

\section{Datasets\label{sec3}}

\begin{table*}[ht]
\centering
\caption{Comparison of 3D SQA datasets. 
\textbf{Source} abbreviations: SCN = ScanNet, 3RS = 3RScan, HME = HM3D, ARK = ARKitScenes, SR+R3D = ScanRefer + ReferIt3D. 
\textbf{Modality} abbreviations: \( S^{(p)} \) = Point Cloud, \( S^{(v)} \) = Video, \( S^{(m)} \) = Multi-view Images. 
\textbf{Suited}: Indicates if the dataset is an Embodied 3D SQA dataset, requiring  the agent to consider \textcolor{black}{its state when answering.}
\textbf{Grounding} denotes whether the dataset includes 3D bounding box annotations for grounding answers to specific objects or regions in the scene.} 
\scriptsize
\resizebox{\textwidth}{!}{%
\begin{tabular}{@{}lllllccc@{}}
\toprule
\textbf{Dataset} & \textbf{Source} & \textbf{Scene} & \textbf{Q\&A} & \textbf{Collection} & \textbf{Modality} & \textbf{Suited} & \textbf{Grounding} \\ \midrule
ScanQA~\cite{azuma2022scanqa} & SCN~\cite{dai2017scannet} & 800 & 41K & Template & \( S^{(p)} \) &$\boldsymbol{\times}$ & \checkmark \\ 
SQA~\cite{ye2021tvcg3dqa} & SCN~\cite{dai2017scannet} & 800 & 6K & Human & \( S^{(p)} \) & $\boldsymbol{\times}$ & $\boldsymbol{\times}$\\ 
FE-3DGQA~\cite{zhao2022toward} & SCN~\cite{dai2017scannet} & 703 & 20K & Human & \( S^{(p)} \) &$\boldsymbol{\times}$ & \checkmark \\ 
CLEVR3D~\cite{yan2023comprehensive} & 3RS~\cite{wald2019rio} & 8,771 & 60K & Template & \( S^{(p)} \) & $\boldsymbol{\times}$& $\boldsymbol{\times}$\\ 
3DVQA~\cite{9866910} & SCN~\cite{dai2017scannet} & 707 & 500K & Template & \( S^{(p)} \) &$\boldsymbol{\times}$ & $\boldsymbol{\times}$\\ 
SQA3D~\cite{ma2022sqa3d} & SCN~\cite{dai2017scannet} & 650 & 33.4K & Human & \( S^{(p)} \) & \checkmark & $\boldsymbol{\times}$\\ 
ScanScribe~\cite{zhu20233d} & SR~\cite{chen2020scanrefer}+R3D~\cite{achlioptas2020referit3d} & 2,995 & 56K & LLM-assisted & \( S^{(p)} \) & \checkmark & $\boldsymbol{\times}$\\ 

\textcolor{black}{HIS-Bench}~\cite{zhao2025his} &  PROX~\cite{hassan2019resolving}+GIMO~\cite{zheng2022gimo} & 31 & 0.5K & LLM-assisted & \(  S^{(p)} \) & \checkmark & $\boldsymbol{\times}$ \\
\textcolor{black}{View2Cap}~\cite{yuan2025empowering}  &  SCN, HM3d~\cite{ramakrishnan2021habitat,yadav2023habitat} & 2841 & 550K & LLM-assisted & \(  S^{(p)} \) & \checkmark & $\boldsymbol{\times}$\\ 

3DMV-VQA~\cite{hong20233d} & HM3d~\cite{ramakrishnan2021habitat} & 5K & 50K & Template & \( S^{(m)} \) & $\boldsymbol{\times}$& $\boldsymbol{\times}$\\ 
OpenEQA~\cite{majumdar2024openeqa} & SCN, HM3d~\cite{ramakrishnan2021habitat,yadav2023habitat} & 180 & 1.6K & Human & \( S^{(m)} \) & $\boldsymbol{\times}$ & $\boldsymbol{\times}$\\ 

Spartun3D~\cite{zhang2024spartun3d} & 3RS~\cite{wald2019rio} & - & 123K & LLM-assisted & \( \{ S^{(m)}, S^{(p)} \} \) & \checkmark & $\boldsymbol{\times}$\\ 
MSQA~\cite{linghu2024multi} & SCN, 3RS, ARK~\cite{baruch2021arkitscenes} & - & 254K & LLM-assisted & \( \{ S^{(m)}, S^{(p)} \} \) & \checkmark & $\boldsymbol{\times}$\\ 
LEO~\cite{huang2023embodied} & SCN+3RS~\cite{wald2019rio} & 3K & 83K & LLM-assisted & \( \{ S^{(m)}, S^{(p)} \} \) & \checkmark & \checkmark \\ 
M3DBench~\cite{li2023m3dbench} & ScanQA~\cite{azuma2022scanqa} & - & 320K & LLM-assisted & \( \{ S^{(m)}, S^{(p)} \} \) & \checkmark & \checkmark \\ 
3D-LLM~\cite{hong20233d3dllm} & Objaverse~\cite{deitke2023objaverse} & - & 300K & LLM-assisted & \( \{ S^{(m)}, S^{(p)} \} \) & \checkmark & \checkmark \\ 
LAMM~\cite{yin2024lamm} & - & - & 186K & LLM-assisted & \( \{ S^{(m)}, S^{(p)} \} \) & \checkmark & \checkmark \\

\textcolor{black}{ROBOSPATIAL}~\cite{song2024robospatial}  &  HOPE~\cite{tyree20226}, GraspNet~\cite{fang1billion}, SCN, etc. & 5k &  3M & LLM-assisted &  \( \{ S^{(m)}, S^{(p)} \} \) & \checkmark & $\boldsymbol{\times}$\\ 

\bottomrule
\end{tabular}%
}
 
\label{table1}
\end{table*}

\textcolor{black}{The growing need for high-level understanding and interaction in 3D environments—such as in robotics and embodied AI—has first stimulated the construction of 3D SQA datasets. 
Datasets are fundamental to model development and evaluation in 3D SQA. 
Existing datasets vary widely in  scene representation, scale, and query complexity.
To provide a systematic overview, this section is organized into three parts. First, \textit{Dataset Structure} examines how scenes and queries are represented, emphasizing the diversity of 3D data modalities and task requirements. Second, \textit{QA Pair Creation} reviews the methodologies for generating question-answer pairs, including template-based pipelines, manual annotations, and LVLM-assisted approaches.
Finally, we conclude this section with \textit{Evaluating LLM-Generated 3D Datasets}, where we analyze emerging trends in dataset development and highlight key characteristics required to support the continued progress of 3D SQA research.}

\subsection{Dataset Structure
}

In the data-driven domain of 3D SQA, structure of datasets significantly influences the scope of the tasks they support. Current datasets differ widely in their representations of 3D scenes, encompassing point clouds, multi-view images, and egocentric perspectives, as well as in the formats of their queries, which range from basic textual inputs to complex multimodal, embodied descriptions. Key dataset attributes such as scale, diversity of modalities, and query complexity significantly influence the design requirements and performance capabilities of 3D SQA models.
Table~\ref{table1} summarizes the key features of existing real-world 3D SQA datasets, providing an overview of their scene representations, query modalities, and scales. 
In Figure~\ref{fig:fig3},  we illustrate the  typical dataset   generation workflow at a higher level of abstraction.

\subsubsection{Scene Modalities and Scale}
\vspace{1mm}
Broadly, the development of 3D SQA datasets has progressed along a timeline evolving from synthetic environments to realistic 3D representations. 

\vspace{1mm}
\noindent\textbf{Synthetic 3D Datasets:}
Due to the initial lack of large-scale real-world 3D point cloud data, early research on 3D Scene Question Answering (3D SQA) relied heavily on synthetic environments to simulate scene-level QA tasks. These environments enabled fully controllable scene construction, semantic labeling, and agent simulation, providing a scalable and flexible foundation for dataset generation.

For instance, EmbodiedQA~\cite{embodiedqa} constructs its dataset by selecting realistic indoor layouts from a subset of SUNCG~\cite{song2017semantic} scenes rendered in the House3D~\cite{wu2018building} simulator. Questions are programmatically generated using predefined functional programs and templates~\cite{johnson2017clevr}, and the corresponding answers are obtained by executing these programs within the virtual environment.Building upon this setup, IQA~\cite{gordon2018iqa} introduced the IQUAD V1 dataset within the AI2-THOR~\cite{kolve2017ai2} simulator. It features 75,000 questions paired with unique scene configurations and focuses on action-conditioned QA, where agents interact with the environment (e.g., opening or picking objects) to complete a task. MP3D-EQA~\cite{wijmans2019embodied} and MT-EQA~\cite{yu2019multi} further incorporated depth maps and multi-target QA tasks, respectively, while remaining confined to synthetic SUNCG~\cite{song2017semantic} scenes.

\vspace{1mm}
\noindent\textbf{Point Cloud Datasets:}
The transition to real-world 3D SQA tasks was marked by the introduction of datasets based on 3D point clouds~\cite{rusu20113d,han2017review}. ScanQA~\cite{azuma2022scanqa} and SQA~\cite{ye2021tvcg3dqa} established foundational benchmarks for this direction. Both datasets were constructed using ScanNet~\cite{dai2017scannet}, with ScanQA generating 41K QA pairs across 800 scenes, and SQA providing 6K manually curated QA pairs with higher linguistic accuracy. Building on these efforts, FE-3DGQA~\cite{zhao2022toward} selected 703 specific scenes from ScanNet and annotated 20K QA pairs, emphasizing foundational QA tasks with dense bounding box annotations to enable spatial grounding.
CLEVR3D~\cite{johnson2017clevr} utilized functional programs and text templates to generate four times the number of questions in ScanQA, introducing a broader range of attributes and question types. Subsequently, 3DVQA~\cite{9866910} expanded on CLEVR3D's framework, leveraging 3D semantic scene graphs and template-based pipelines to generate questions and answers. By selecting 707 scenes, 3DVQA produced 500K QA pairs, significantly enriching task diversity and complexity. Similarly, SQA3D~\cite{ma2022sqa3d} marked a significant advancement in agent-centric 3D QA. It curated 33.4K manually annotated QA pairs across 650 scenes, focusing on linking queries to  agent position and orientation. 
This dataset enabled deeper exploration of tasks that integrate agent perspectives with spatial understanding. \textcolor{black}{HIS-Bench~\cite{zhao2025his} is the first benchmark specifically tailored for Human-In-Scene (HIS)~\cite{araujo2023circle,hassan2021stochastic} understanding. As a small yet high-quality dataset, it focuses on question answering tasks that center around human–agent interactions within 3D scenes. In contrast, View2Cap~\cite{yuan2025empowering} emphasizes spatially grounded reasoning and constructs a large-scale corpus of 550K QA pairs, making it one of the most representative benchmarks for evaluating whether models can effectively understand 3D positional information.}

\vspace{1mm}
\noindent\textbf{Multi-View Datasets:}
To better align with human perception, multi-view datasets have been introduced, focusing on reasoning across different perspectives rather than relying solely on single point cloud representations. In this direction, 3DMV-VQA~\cite{hong20233d} includes 5K scenes from the HM3D dataset~\cite{ramakrishnan2021habitat}, generating 50K QA pairs. The images are rendered using the Habitat framework~\cite{ramakrishnan2021habitat,savva2019habitat,szot2021habitat}, emphasizing multi-view reasoning. On the other hand, OpenEQA~\cite{majumdar2024openeqa} not only selects scenes from HM3D but also incorporates Gibson~\cite{xia2018gibson} and ScanNet~\cite{dai2017scannet}, ultimately choosing 180 high-quality scenes with 1.6K QA pairs. Unlike other datasets, it prioritizes quality over scale, making it a significant contribution to high-quality 3D SQA benchmarks.

\vspace{1mm}
\noindent\textbf{Multimodal  Datasets:}
Recent advances in 3D SQA datasets emphasize integrating point clouds, images, and textual data to form rich multimodal representations. These approaches aim to capture spatial, semantic, and contextual cues for more comprehensive scene understanding. A notable example is Spartun3D~\cite{zhang2024spartun3d}, which selects scenes from 3RScan~\cite{wald2019rio} and generates 123K QA pairs focused on situational tasks. Similarly, MSQA~\cite{linghu2024multi} builds 254K QA pairs from multimodal datasets~\cite{dai2017scannet, wald2019rio, baruch2021arkitscenes}, using point clouds and object images as inputs to better align with real-world embodied intelligence scenarios.  
With the popularity of LLMs, instruction tuning datasets have also emerged as an important extension of multimodal datasets, enhancing the generalization capabilities of 3D SQA models by aligning 3D data with textual descriptions. For instance, ScanScribe~\cite{zhu20233d} collects RGB-D scans of indoor scenes from ScanNet and 3R-Scan, incorporating diverse object instances from Objaverse~\cite{deitke2023objaverse}. It uses QA pairs from ScanQA and referential expressions from ScanRefer~\cite{chen2020scanrefer} and ReferIt3D~\cite{achlioptas2020referit3d}, generating 56.1K object instances from 2,995 scenes through templates and GPT-3~\cite{brown2020language}. Similarly, LEO~\cite{huang2023embodied} constructs 83K 3D-text pairs by collecting captions at object, object-in-scene, and scene levels~\cite{luo2024scalable, achlioptas2020referit3d, zhu20233d, chen2021scan2cap}. \textcolor{black}{Robospatial~\cite{song2024robospatial} features real-world indoor and tabletop scenes captured in the form of 3D scans and egocentric images, annotated with rich spatial information relevant to robotic tasks. The dataset includes 1 million images, 5,000 3D scans, and 3 million annotated spatial relation graphs and QA pairs, making it one of the largest and most comprehensive benchmarks available to date.
}

Along similar lines, M3DBench~\cite{li2023m3dbench} leverages multiple existing 
and LLMs to generate 320K instruction-response pairs, enriching multimodal 3D data for a wide range of 3D-language tasks. 3D-LLM~\cite{hong20233d3dllm}  creates over 300K 3D-text pairs using assets like Objaverse, ScanNet, and HM3D, while LAMM~\cite{yin2024lamm} employs GPT-API and self-instruction methods~\cite{wang2022self} to produce 186K language-image pairs and 10K language-3D pairs. 
\subsubsection{Query Modalities and Complexity}

\begin{figure}[t]
    \centering
    \includegraphics[width=\linewidth]{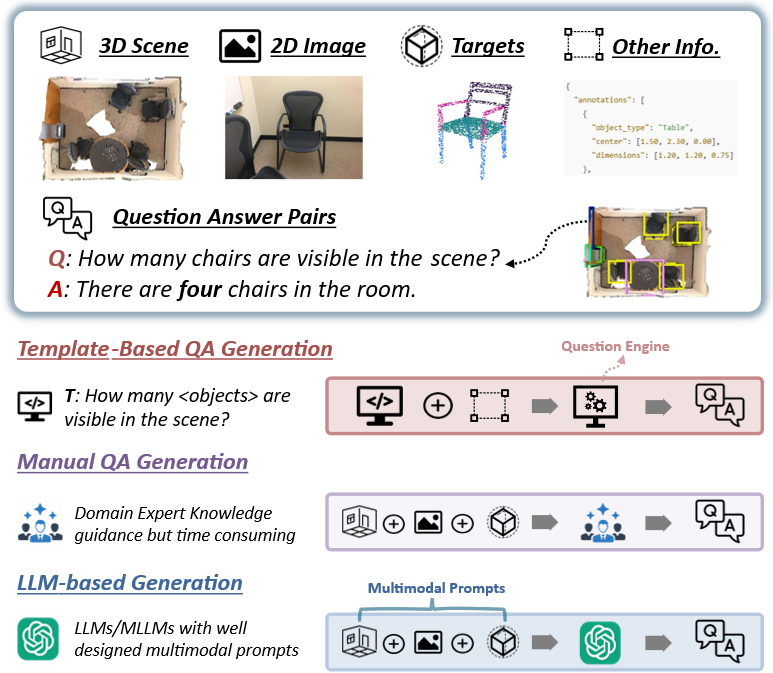}
    \caption{Dataset generation workflow.}
    \label{fig:fig3}
  
\end{figure} 

In 3D SQA, a query represents the input question or prompt that, when paired with a 3D scene, guides the task of providing an answer. Over time, query modalities in 3D SQA have evolved from simple text-based inputs to more complex, multimodal, and agent-centric formats. Here, we summarise the datasets from the query modality perspective,  which is a critical consideration for dataset selection in performance evaluation.  


\vspace{1mm}
\noindent\textbf{Basic Text Queries:}
Early 3D SQA datasets primarily employed straightforward text-based queries that focused on scene-level attributes, such as object counting or identification. These datasets aimed to evaluate foundational 3D scene understanding, often without considering the agent's position, interaction, or perspective within the environment. For example, datasets like {ScanQA}~\cite{azuma2022scanqa} and {SQA}~\cite{ye2021tvcg3dqa} feature questions such as \textit{"How many chairs are in the room?"}.
Such purely textual questions  fail to capture  complex embodied scenarios as they lack description of an agent's spatial or contextual relationship with the scene.
Consequently, these datasets 
are limited in scope, as reflected in Table~\ref{table1}, where the lack of \textit{Suited} queries indicates their omission of agent-centric contexts. This limitation underscores the evolution toward richer, more contextualized datasets in the later  3D SQA research.

\vspace{1mm}
\noindent\textbf{Agent-Centric Text Queries:} 
The introduction of agent-centric descriptions marked a significant shift in query complexity. {SQA3D}~\cite{ma2022sqa3d} was one of the first datasets to incorporate contextualized questions, where textual queries were enhanced with references to the agent’s position or orientation. In this case, a typical query might describe the agent's location, such as \textit{"Sitting at the edge of the bed and facing the couch."}. We mark datasets enabling such queries as \textit{Suited} in Table~\ref{table1}. 

\vspace{1mm}
\noindent\textbf{Multimodal Agent-Centric Queries:} 
Recently, {SPARTUN3D}~\cite{zhang2024spartun3d} and {MSQA}~\cite{linghu2024multi} introduced richer spatial descriptions and multimodal query inputs.
The former provided  detailed spatial information, enabling queries such as \textit{"You are standing beside a trash bin while there is a toilet in front of you."}. 
Similarly, MSQA integrated textual descriptions, explicit spatial coordinates, and agent orientation in the queries. Additionally, first-person view images were included.
These multimodal approaches enable more realistic scenarios by combining spatial, visual, and linguistic contexts.

\vspace{1mm}
\noindent\textbf{Instruction-Tuned Queries:} 
Recent datasets, such as ScanScribe~\cite{zhu20233d}, LEO~\cite{huang2023embodied}, and M3DBench~\cite{li2023m3dbench}, have also expanded query modalities further to support instruction tuning tasks. They leverage agent-centric queries enriched with multimodal inputs, such as spatially grounded textual descriptions and multimodal instructions.
For example, LEO incorporates multimodal instructions to fine-tune models for agent tasks like real-time navigation or object interaction.
M3DBench focuses on generalization across diverse real-world tasks by utilizing rich multimodal data.
These instruction-tuning datasets ensure models are well-equipped to address practical, real-world tasks by aligning textual instructions with spatial and visual contexts.


\begin{table*}[ht]
\centering
\caption{Examples of 3D SQA tasks, identified by their objectives, along with  representative example questions. The tasks cover a range of capabilities, including object identification, spatial reasoning, attribute querying, multi-hop reasoning, and planning. These tasks demonstrate the diverse applications and challenges addressed in 3D SQA, requiring models to integrate spatial, semantic, and task-specific understanding.}
\small
\begin{tabular}{@{}p{0.3\textwidth}p{0.7\textwidth}@{}}
\toprule
\textbf{Task} & \textbf{Example Question} \\ \midrule
Object Identification & What is the object next to the red chair in the room? \\ \midrule
Spatial Reasoning & Where is the table located relative to the sofa? \\ \midrule
Attribute Querying & What is the color of the sphere on the shelf? \\ \midrule
Object Counting & How many chairs are there in the room? \\ \midrule
Attribute Comparison & Which is taller, the lamp or the bookshelf? \\ \midrule
Multi-hop Reasoning & Find the green bottle in the kitchen. What is on the shelf above it? \\ \midrule
Navigation & Guide the agent to the bedroom and locate the bedside table. \\ \midrule
Robotic Manipulation & Pick up the blue block and place it on the red cube. \\ \midrule
Object Affordance & What can be done with the knife on the counter? \\ \midrule
Functional Reasoning & How would you use the tools in the box to fix the broken chair? \\ \midrule
Multi-round Dialogue & User: Where is the TV? \newline Model: It is in the living room on the wall. \newline User: What is under the TV? \\ \midrule
Planning & Plan a sequence of actions to make a cup of tea using objects in the kitchen. \\ \midrule
Task Decomposition & Break down the task of assembling a desk into individual steps. \\
\bottomrule
\end{tabular}
\label{app_table_q_examples}
\end{table*}

\subsection{ QA Pair Creation}
\textcolor{black}{Beyond the representation of multimodal scenes, a fundamental aspect of 3D SQA lies in the nature of the questions being asked. The design and generation of  QA pairs directly define the scope, complexity, and semantic depth of the task.}
Early datasets relied on manual annotation, while recent efforts have adopted templates and LVLMs to improve scalability and diversity. \textcolor{black}{These advances have enabled datasets to include a wider range of question types, from object identification to spatial relationships and task-specific queries.} 

\subsubsection{Methods for QA Pair Generation}
QA pair generation in 3D SQA datasets balances between manual annotation, template-based pipelines, and LLM-assisted methods. Manual annotation ensures high-quality and contextual accuracy, while template-based approaches enable scalable generation with logical consistency. Recently, LLMs have further automated the process, enabling diverse multimodal  QA pairs at scale. This progression, also  apparent in Figure~\ref{fig:fig3}, reflects the evolution of dataset creation techniques.

\vspace{1mm}
\noindent\textbf{Template-Based Generation:} 
Template-based generation was introduced as an early solution to scale up QA pair creation while maintaining structural consistency. This approach typically relies on predefined syntactic patterns combined with scene annotations such as object labels, locations, and relationships. For instance, ScanQA~\cite{azuma2022scanqa} leveraged a T5-based model~\cite{raffel2020exploring} to generate seed questions from the ScanRefer~\cite{chen2020scanrefer} dataset, followed by human refinement to ensure naturalness and diversity.
In parallel, datasets such as CLEVR3D~\cite{yan2023comprehensive}, 3DVQA~\cite{9866910}, and 3DMV-VQA~\cite{hong20233d} adopted programmatic question generation grounded on 3D semantic scene graphs. These structures capture object relationships and scene layouts, enabling systematic synthesis of questions involving spatial reasoning, attribute comparison, and multi-step logic.

While template-based methods significantly improve scalability and logical consistency, they often suffer from limited linguistic diversity and contextual richness. As a result, generated questions may become repetitive, overly generic, or detached from the agent's embodied perspective.

\vspace{1mm}
\noindent\textbf{Manual Annotation:}
Researchers have also pursued manual annotation to address the limitations of template-based methods. Manual approaches prioritize linguistic precision and contextual relevance, creating datasets that are smaller in scale but of higher quality. For instance, SQA~\cite{ye2021tvcg3dqa} curated 6K QA pairs with an emphasis on linguistic accuracy, while FE-3DGQA~\cite{zhao2022toward} selected 703 scenes from ScanNet~\cite{dai2017scannet} and annotated 20K QA pairs, grounding answers with bounding box annotations. Similarly, OpenEQA~\cite{majumdar2024openeqa} curated 1.6K QA pairs from 180 high-quality scenes. SQA3D~\cite{ma2022sqa3d} contributed 33.4K QA pairs across 650 scenes, tailored specifically for agent-centric tasks. Despite their time-intensive nature, fully curated datasets play a critical role in ensuring accuracy and contextual alignment, complementing the template-based methods.

\vspace{1mm}
\noindent\textbf{LLM-Assisted Generation:}
Recent methods have increasingly leveraged LLMs to automate the generation of QA pairs, enhancing both scalability and diversity. Notable examples include Spartun3D~\cite{zhang2024spartun3d} and MSQA~\cite{linghu2024multi}, both of which utilize scene graphs to structure spatial and semantic relationships. Spartun3D employs GPT-3.5 to generate agent-centric questions, emphasizing situated reasoning and exploration, resulting in 123K QA pairs. MSQA takes a similar approach with GPT-4V, focusing on situated QA generation guided by semantic scene graphs, producing 254K QA pairs.
These datasets highlight how integrating LLMs with scene graphs facilitates the creation of rich and contextually relevant QA pairs while maintaining scalability.

Additionally, LLMs have been instrumental in constructing instruction tuning datasets to improve model generalization across diverse multimodal tasks. ScanScribe~\cite{zhu20233d} utilizes GPT-3 to transform ScanRefer annotations into scene descriptions using template-based refinement.
LEO~\cite{huang2023embodied} adopts GPT-4 with Object-centric Chain-of-Thought (O-CoT) prompting to ensure logical consistency, resulting in 83K object- and scene-level 3D-text pairs. 
M3DBench~\cite{li2023m3dbench} and 3D-LLM~\cite{hong20233d3dllm} use GPT-4 to create multimodal prompts based on object attributes and scene-level inputs, generating 320K and 300K instruction-response pairs, respectively. 
Together, these datasets demonstrate the growing role of LLMs in automating the generation of high-quality, multimodal data for 3D SQA, laying the foundation for models capable of handling complex embodied intelligence tasks.

\subsubsection{Question Design in 3D SQA}
While the previous section focused on how QA pairs are generated, this section turns to the content of the questions themselves, which fundamentally defines the nature and capability of 3D SQA tasks.
With advancements in language and vision modelling, 3D SQA questions have evolved along several dimensions: from simple to complex tasks, non-situated to situated contexts, and static to dynamic scenarios. 
\textcolor{black}{To exemplify the nature of these questions, we enlist the common 3D SQA tasks and representative question in  Table~\ref{app_table_q_examples}.}

\vspace{1mm}
\noindent\textbf{Task Complexity - From Basic to Advanced:}
3D SQA covers a diverse spectrum of question tasks designed to assess models’ understanding of 3D environments and their reasoning abilities.
Basic tasks, such as object identification, spatial reasoning, attribute querying, object counting, and attribute comparison, are featured in datasets like SQA~\cite{ye2021tvcg3dqa}, ScanQA~\cite{azuma2022scanqa}, FE-3DGQA~\cite{zhao2022toward}, 3DVQA~\cite{9866910} and CLEVR3D~\cite{yan2023comprehensive}. Among these, FE-3DGQA introduced more complex, free-form questions that require models not only to ground answer-relevant objects but also to identify contextual relationships between them. Similarly, CLEVR3D  emphasized relational reasoning by incorporating questions that integrate objects, attributes, and their interrelationships, pushing models further to handle intricate contextual dependencies.

As 3D SQA evolves, tasks demanding a deeper understanding of spatial and visual context have emerged, challenging models to engage with dynamic and context-aware reasoning. These tasks include multi-hop reasoning (SQA3D~\cite{ma2022sqa3d}), navigation (SQA3D~\cite{ma2022sqa3d}, LEO~\cite{huang2023embodied}, 3D-LLM~\cite{hong20233d3dllm}, M3DBench~\cite{li2023m3dbench}, MSQA~\cite{linghu2024multi}), robotic manipulation (LEO), object affordance (Spartun3D~\cite{zhang2024spartun3d}), functional reasoning (OpenEQA~\cite{majumdar2024openeqa}), multi-round dialogue (LEO, M3DBench, 3D-LLM), planning (LEO, M3DBench, Spartun3D), and task decomposition (3D-LLM). These advanced tasks challenge models to dynamically reason and navigate complex 3D environments while capturing intricate spatial and relational details. Notably, OpenEQA~\cite{majumdar2024openeqa} stands out as the first open-vocabulary dataset for embodied question answering.

\vspace{1mm}
\noindent\textbf{Situated vs. Non-Situated Questions:}
Based on the required level of interaction and contextual understanding, 3D VQA questions can be categorized into  situated and non-situation types. The latter focus on static reasoning, testing a model’s ability to interpret spatial relationships, attributes, and object properties within fixed 3D scenes. Datasets like SQA~\cite{ye2021tvcg3dqa}, ScanQA~\cite{azuma2022scanqa}, FE-3DGQA~\cite{zhao2022toward}, 3DVQA~\cite{9866910}, CLEVR3D~\cite{yan2023comprehensive}, and LAMM~\cite{yin2024lamm} primarily include non-situated questions that evaluate understanding within static spatial contexts.

Conversely, situated questions involve dynamic reasoning, requiring interaction with the 3D environment and comprehension of contextual or sequential information. These questions test models’ ability to navigate, plan, and adapt to dynamic scenarios and often include temporal or embodied elements. Situated questions appear in datasets like SQA3D~\cite{ma2022sqa3d}, LEO~\cite{huang2023embodied}, 3D-LLM~\cite{hong20233d3dllm}, M3DBench~\cite{li2023m3dbench}, MSQA~\cite{linghu2024multi}, Spartun3D~\cite{zhang2024spartun3d}, 3DMV-VQA~\cite{hong20233d}, and OpenEQA~\cite{majumdar2024openeqa}. This categorization enables a comprehensive evaluation of 3D VQA systems. 

\vspace{1mm}
\noindent\textbf{Temporal Aspect in 3D SQA:}
Most 3D SQA datasets limit questions to a single time slot, reflecting the static nature of the environments they evaluate. This restriction simplifies reasoning by focusing on a specific moment within the 3D scene. However, datasets like OpenEQA~\cite{majumdar2024openeqa} now introduce dynamic scenarios that allow for multiple time slots, enabling tasks that require episodic memory and active exploration. This temporal dimension challenges models to integrate sequential information and represents a significant step forward for advancing 3D SQA.

\subsection{Evaluating LLM-Generated 3D Datasets}
While   LLM adoption has significantly advanced 3D SQA datasets, 
ensuring their quality, reliability, and practical utility  remains an open  challenge. 
Current evaluation methods primarily rely on manual assessments. For example, LEO~\cite{huang2023embodied} evaluates QA pairs through expert review, reporting metrics like overall accuracy and contextual relevance. MSQA~\cite{linghu2024multi} adopts a comparative approach, sampling QA pairs from its dataset and comparing them against a benchmark dataset such as SQA3D~\cite{ma2022sqa3d}, with scores based on contextual accuracy, factual correctness, and overall quality. Similarly, Spartun3D~\cite{zhang2024spartun3d} employs expert validation by randomly sampling instances to ensure that the generated data meets expected quality standards. These manual evaluations provide valuable insights into dataset quality but face limitations in scalability, labour intensity, and subjectivity.

To address these limitations, automated evaluation frameworks are currently  needed. Potential solutions include embedding-based metrics for semantic alignment, logical consistency checks for QA coherence, and task-specific metrics for spatial accuracy and multimodal integration. 

\section{Evaluation Metrics\label{sec4}}

Standardized  evaluation metrics are crucial to gauge advances in  3D SQA and ensure dataset suitability for downstream tasks. 
Contemporary 3D SQA literature either  uses traditional or LLM-based metrics for the evaluation purpose.

\vspace{1mm}
\subsection{Traditional metrics}  
\vspace{1mm}
\textcolor{black}{3D SQA methods often adopt standard language-based evaluation metrics to assess the correctness and relevance of predicted answers. Commonly used metrics include:}

- \textit{Exact Match (EM@1, EM@10)} evaluates whether the predicted answer exactly matches one of the ground truth answers, either in top-1 or top-10 predictions. It is a strict metric that captures answer accuracy but does not tolerate synonyms, paraphrasing, or minor variations in wording.

- \textit{BLEU}~\cite{papineni2002bleu} measures n-gram precision by computing how many overlapping word sequences of length 1 to 4 exist between the prediction and reference. Although widely used in machine translation, BLEU may penalize semantically correct but lexically different responses.

- \textit{ROUGE-L}~\cite{lin2004rouge} computes the length of the longest common subsequence between the predicted and reference texts. It captures sentence-level similarity and is more tolerant to word order variations than BLEU.

- \textit{METEOR}~\cite{banerjee2005meteor} aligns predicted and reference texts using stemming, synonym matching (via WordNet~\cite{miller1995wordnet}), and weighted precision/recall. It is more sensitive to semantic similarity than BLEU or ROUGE, making it suitable for evaluating answer variants.

- \textit{CIDEr}~\cite{vedantam2015cider} measures the consensus of a generated answer across multiple references by computing TF-IDF weighted n-gram similarity. It emphasizes content relevance based on how commonly certain terms appear in ground truth responses.

- \textit{SPICE}~\cite{anderson2016spice} parses both prediction and reference into scene graphs, and compares semantic structures such as objects, attributes, and relationships. It is designed to align more closely with human judgment in visual tasks, making it potentially useful in spatially grounded QA.

These metrics were first adopted in ScanQA~\cite{azuma2022scanqa} and have since been widely used in other 3D SQA benchmarks such as CLEVR3D~\cite{yan2023comprehensive}, 3DGQA~\cite{zhao2022toward}, and ScanScribe~\cite{zhu20233d}. While effective for evaluating linguistic accuracy and surface-level fluency, these traditional metrics often fall short in capturing the deeper contextual reasoning, spatial understanding, and embodied interactions central to 3D SQA tasks.

\begin{table*}[htbp]
\caption{Overview of techniques for 3D SQA. Methods are categorized as Task-Specific (T-S), Pretraining-Based (P-B) and Zero-Shot (Z-S).  \textcolor{black}{P-B (w I-T) denotes Pretraining-Based methods further enhanced with Instruction Tuning to better adapt to task-specific instructions}. Scene modalities are represented as \( S^{(p)} \) for Point Cloud, \( S^{(m)} \) for Image, and \( \{ S^{(m)}, S^{(p)} \} \) for Multimodal.}
\centering
\small
\begin{tabular}{l|ccccc}
\toprule
\textbf{Method} & \textbf{Type} & \textbf{Scene Modality} & \textbf{Scene Encoder} & \textbf{Text Encoder} & \textbf{Answer Module} \\ 
\midrule
ScanQA~\cite{azuma2022scanqa} & T-S & \( S^{(p)} \) & VoteNet~\cite{qi2019deep} & BiLSTM~\cite{graves2012long} & MLP \\ 
3DQA-TR~\cite{ye2021tvcg3dqa} & T-S & \( S^{(p)} \) & Group-Free~\cite{liu2021group} & BERT~\cite{kenton2019bert} & MLP \\ 
TransVQA3D~\cite{yan2023comprehensive} & T-S & \( S^{(p)} \) & PointNet++~\cite{qi2017pointnet++} & BERT~\cite{kenton2019bert} & MLP \\ 
FE-3DGQA~\cite{zhao2022toward} & T-S & \( S^{(p)} \) & PointNet++~\cite{qi2017pointnet++} & T5~\cite{raffel2020exploring} & Linear Layer \\ 
SIG3D~\cite{man2024situational} & T-S & \( S^{(p)} \) &  OpenScene~\cite{peng2023openscene} & BiLSTM~\cite{graves2012long} & MLP \\ 
3D-CLR~\cite{hong20233d} & T-S & \( S^{(m)} \) & CLIP-LSeg~\cite{li2022language} & CLIP~\cite{radford2021learning} & 3D CNN \\ 

BridgeQA~\cite{mo2024bridging} & T-S &  \( \{ S^{(m)}, S^{(p)} \} \) & VoteNet\&BLIP~\cite{li2023blip} & BLIP~\cite{li2023blip} & Transformer \\
\hline
3DVLP~\cite{zhang2024vision} & P-B & \( S^{(p)} \) & PointNet++~\cite{qi2017pointnet++} & CLIP~\cite{radford2021learning} & MLP \\ 
CLIP-Guided~\cite{parelli2023clip} & P-B & \( S^{(p)} \) & VoteNet\&Transformer~\cite{vaswani2017attention} & CLIP~\cite{radford2021learning} & MLp \\ 
Multi-CLIP~\cite{delitzas2023multi} & P-B & \( S^{(p)} \) & VoteNet\&Transformer~\cite{vaswani2017attention} & CLIP~\cite{radford2021learning} & MLP \\ 
3D-VisTA~\cite{zhu20233d} & P-B & \( S^{(p)} \) & PointNet++~\cite{qi2017pointnet++} & BERT & MLP \\ 
GPS~\cite{jia2025sceneverse} & P-B & \( S^{(p)} \) & PointNet++~\cite{qi2017pointnet++} & Transformer~\cite{vaswani2017attention} & Transformer \\ 
LM4Vision~\cite{pang2023frozen} & P-B(w I-T) & \( S^{(p)} \) & VoteNet~\cite{qi2019deep} &  LSTM~\cite{hochreiter1997long} & LLaMA~\cite{touvron2023llama}  \\ 
3D-LLM~\cite{hong20233d3dllm} & P-B(w I-T) & \( S^{(m)} \) & BLIP2~\cite{li2023blip}  & BLIP2~\cite{li2023blip} & BLIP2~\cite{li2023blip} \\ 
LEO~\cite{huang2023embodied} & P-B(w I-T) & \( S^{(p)} \) & PointNet++\& ST~\cite{chen2022language} & ConvNext~\cite{liu2022convnet} & Vicuna~\cite{chiang2023vicuna} \\ 
LAMM~\cite{yin2024lamm} & P-B(w I-T) & \( S^{(p)} \) & PointNet++~\cite{qi2017pointnet++} & SentencePiece~\cite{kudo2018sentencepiece} & Vicuna~\cite{chiang2023vicuna} \\ 
M3DBench~\cite{li2023m3dbench} & P-B(w I-T) & \( S^{(p)} \) & PointNet++\& Transformer & Opt~\cite{zhang2022opt} & Opt~\cite{zhang2022opt} \\ 
\textcolor{black}{LL3DA} ~\cite{chen2024ll3da}& P-B(w I-T) & \( S^{(p)} \) & Vote2Cap-DETR~\cite{chen2023end} &  Opt~\cite{zhang2022opt} &  Opt~\cite{zhang2022opt} \\ 
HIS-GPT~\cite{zhao2025his} & P-B(w I-T) & \( S^{(p)} \) & Uni3d~\cite{zhou2023uni3d}& Vicuna~\cite{chiang2023vicuna} & Vicuna~\cite{chiang2023vicuna} \\ 
\textcolor{black}{View2Cap}~\cite{yuan2025empowering} & P-B(w I-T) & \( S^{(p)} \) & Uni3d~\cite{zhou2023uni3d}& LLaMa~\cite{touvron2023llama} & LLaMa~\cite{touvron2023llama}\\ 
\textcolor{black}{SplatTalk} ~\cite{thai2025splattalk} &  P-B(w I-T) & \( S^{(m)} \) &3DGS~\cite{qin2024langsplat}  & LLaVA-OV~\cite{thai2025splattalk} & LLaVA-OV~\cite{thai2025splattalk} \\ 
\textcolor{black}{Scene-LLM}~\cite{fu2024scene} & P-B(w I-T) & \( S^{(p)} \) & PVCNN~\cite{liu2019point} & Llama~\cite{touvron2023llama} & Llama~\cite{touvron2023llama}\\ 
\textcolor{black}{Chat-Scene}~\cite{huang2024chat} & P-B(w I-T) & \( \{ S^{(m)}, S^{(p)} \} \) & Mask3d~\cite{schult2023mask3d}\& Uni3d~\cite{zhou2023uni3d} & Vicuna~\cite{chiang2023vicuna} & Vicuna~\cite{chiang2023vicuna}\\ 

\hline
SQA3D~\cite{ma2022sqa3d} & Z-S & \( S^{(p)} \) & Scan2Cap~\cite{chen2021scan2cap} & GPT-3~\cite{brown2020language} & GPT-3 \\ 
LAMM~\cite{yin2024lamm} & Z-S & \( S^{(p)} \) & PointNet++~\cite{qi2017pointnet++} & SentencePiece~\cite{kudo2018sentencepiece} & Vicuna \\ 
EZSG~\cite{singh2024evaluating} & Z-S & \( S^{(m)} \) & GPT-4V~\cite{yang2023dawn} & GPT-4V~\cite{yang2023dawn} & GPT-4V \\ 
OpenEQA~\cite{majumdar2024openeqa} & Z-S & \( S^{(m)} \) & GPT-4V~\cite{yang2023dawn} & GPT-4V~\cite{yang2023dawn} & GPT-4V \\ 
MSQA~\cite{linghu2024multi} & Z-S & \( S^{(m)} \) & GPT-4o~\cite{achiam2023gpt} & GPT-4o~\cite{achiam2023gpt} & GPT-4o \\ 
LEO~\cite{huang2023embodied} & Z-S &  \( \{ S^{(m)}, S^{(p)} \} \) & PointNet++\& ST~\cite{chen2022language} & ConvNext~\cite{liu2022convnet} & Vicuna~\cite{chiang2023vicuna} \\  
Spartun3D-LLM~\cite{zhang2024spartun3d} & Z-S &  \( \{ S^{(m)}, S^{(p)} \} \) & PointNet++~\cite{qi2017pointnet++} & CLIP~\cite{radford2021learning} & Vicuna \\ 

\toprule
\end{tabular}
\label{table:methods}

\end{table*}

\vspace{1mm}
\subsection{LLM-based metrics}  
\vspace{1mm}

While traditional metrics such as Exact Match, BLEU, and ROUGE provide efficient evaluations of linguistic overlap, they often fail to capture the deeper semantic correctness and contextual alignment required in 3D SQA tasks. For example, given the question “What is on the table in front of the couch?”, an answer of “a lamp” and a reference answer of “the lamp” would receive a score of 0 under Exact Match, despite being semantically equivalent. Conversely, an incorrect answer like “a chair” may still receive a relatively high BLEU score due to n-gram overlap. These limitations highlight the need for more semantically aware evaluation methods.

To address this, recent works have proposed using large language models (LLMs) to directly assess the quality of model outputs. A representative approach is \textit{LLM-Match}~\cite{majumdar2024openeqa}, which leverages GPT to score the semantic correctness of predicted answers in open-ended settings. Given a question \( Q_i \), a ground-truth reference answer \( T_i^* \), and a model-generated answer \( T_i \), the LLM is prompted to assign a relevance score \( \sigma_i \in \{1, 2, 3, 4, 5\} \), where 1 indicates incorrect and 5 indicates fully correct. The final correctness score is normalized as:

\begin{equation}
C = \frac{1}{N} \sum_{i=1}^{N} \frac{\sigma_i - 1}{4} \times 100\%
\end{equation}

This approach enables a more fine-grained, human-aligned evaluation of semantic similarity and contextual appropriateness. Compared to rule-based metrics, LLM-based methods are better equipped to handle open-vocabulary answers, embodied references, and spatially grounded reasoning—key characteristics of 3D SQA tasks. Similarly, MSQA~\cite{linghu2024multi} uses GPT to assess the quality of answers based on nuanced reasoning, aligning them with contextual expectations. Compared to traditional metrics, LLM-based methods currently excel at simulating real-world reasoning and capturing semantic subtleties, making them particularly valuable for evaluating complex multimodal tasks.

In summary, traditional metrics provide a strong foundation for evaluating linguistic and structural quality, while LLM-based metrics offer deeper insights into contextual alignment and reasoning. Combining the complementary properties of these metrics can offer a comprehensive framework for assessing 3D SQA performance. 

\begin{figure}[t]
    \centering
    \includegraphics[width=0.95\linewidth]{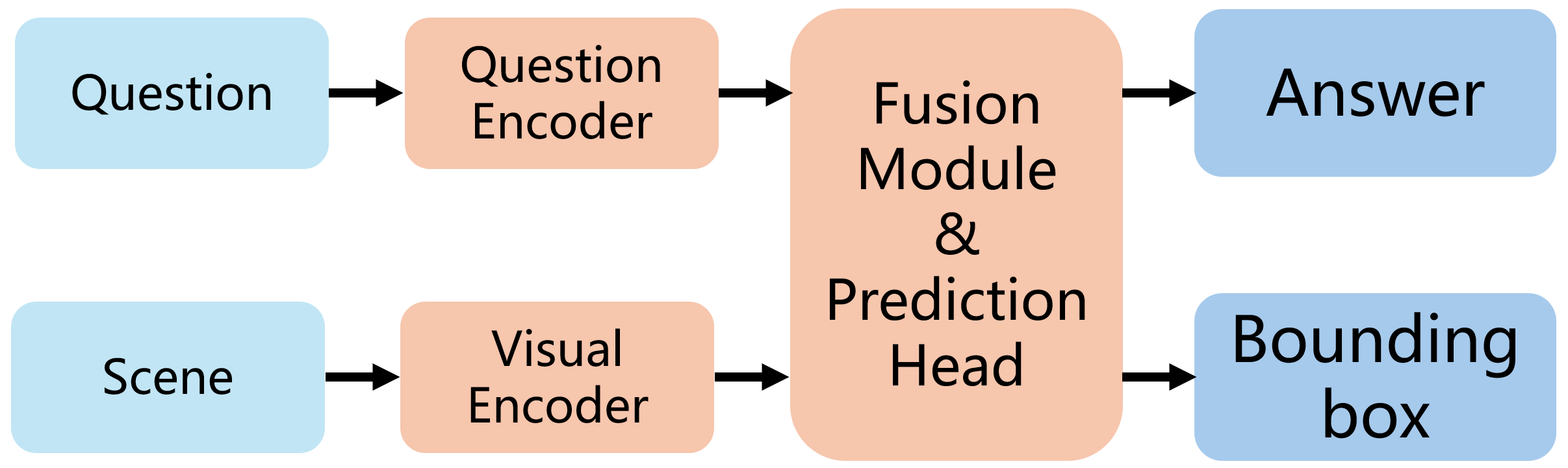}
  \caption{\textcolor{black}{Overview of a generalized 3D SQA pipeline. The scene input—represented as images or point clouds—is processed by a visual encoder, while the question input—comprising textual and potentially egocentric visual components—is encoded separately. The resulting features are fused via a dedicated fusion module and passed to a joint prediction head that outputs the answer and optional 3D bounding boxes. Recent approaches enhance this pipeline by incorporating large vision-language models (LVLMs) to support instruction tuning and zero-shot reasoning.}}
    \label{fig:fig4}
\end{figure}

\section{Taxonomy of 3D SQA Methods
\label{sec5}}
With the progressive development of benchmark datasets and evaluation protocols, the 3D SQA task has become well-defined in terms of input-output formats, annotation granularity, and quality criteria. These foundations have in turn enabled the emergence of diverse modeling paradigms aiming to bridge vision, language, and spatial understanding.

3D SQA methods can be categorized into three primary types, as shown in Table~\ref{table:methods}. 
i) \textit{Task-Specific Methods} rely on predefined answers and specialized architectures to address specific tasks.
ii)~\textit{Pretraining-Based Methods}  leverage large-scale datasets to align multimodal representations and fine-tune for task-specific objectives.
iii) \textit{Zero-Shot Learning Methods} also utilize pretrained LLMs and VLMs to generalize to new tasks, albeit without additional fine-tuning.

These categories reflect the field’s evolution toward more flexible, scalable solutions. Despite their differences, our survey reveals that most existing approaches share a similar architectural design pattern. As illustrated in Figure~\ref{fig:fig4}, they typically follow a unified pipeline involving modality-specific encoders, a fusion module, and prediction heads for answer generation and optional spatial grounding. This abstraction provides a useful framework for analyzing and comparing different methods under a common lens.

\begin{figure*}[t]
    \centering
\includegraphics[width=0.90\linewidth]{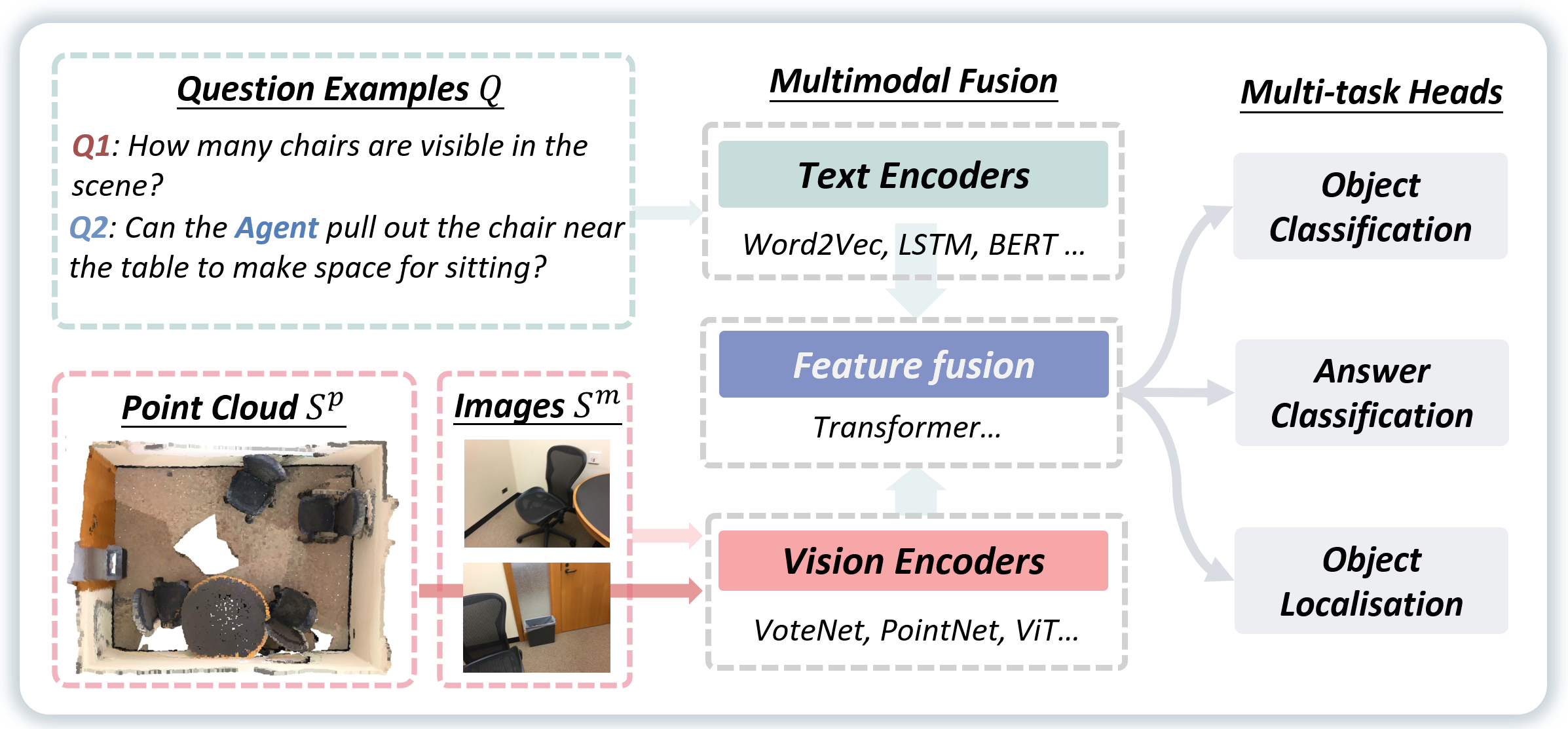}
  \caption{\textcolor{black}{Typical architecture of task-specific 3D SQA methods. Scene and query (question) features are encoded separately, fused via a transformer-based module, and used to predict the answer, optionally with bounding boxes and object categories.}}
    \label{scanqa}
\end{figure*}

\subsection{Task-Specific Methods}
These methods are designed for specific tasks using  closed-set classification approach. As illustrated in Figure~\ref{scanqa},
\textcolor{black}{task-specific methods employ designated vision and language encoders to extract features from the scene and the question, respectively. These features are then fused—typically using a Transformer~\cite{vaswani2017attention} or its variants—and the question-answering task is formulated as a classification problem to predict the final answer. Some methods further incorporate optional prediction of object bounding boxes and categories relevant to the query.}

\vspace{1mm}
\noindent\textbf{Point Cloud Methods:}
Early 3D SQA methods designed specifically for point cloud inputs typically adopt a modular pipeline consisting of scene encoding, query encoding, multimodal fusion, and answer prediction. A representative example is ScanQA~\cite{azuma2022scanqa}, which represents the 3D scene as a point cloud and processes it using VoteNet~\cite{qi2019deep} and PointNet++~\cite{qi2017pointnet++} to extract spatial and instance-level features. The textual question is encoded using GloVe~\cite{pennington2014glove} embeddings followed by a BiLSTM~\cite{graves2012long} to capture contextual semantics. The visual and language features are fused via transformer-based cross-modal attention modules, and the final answer is predicted from a closed vocabulary using a classification head. \textcolor{black}{This pipeline characterizes the early paradigm of 3D SQA, where task-specific models operate under closed-set settings by employing dedicated encoders for different scene modalities (e.g., RGB, RGB-D), while maintaining a consistent architecture for feature fusion and answer prediction.}


Building on this foundation, later methods introduced more sophisticated encoders and fusion strategies. For example, 3DQA-TR~\cite{ye2021tvcg3dqa} replaced VoteNet with Group-Free~\cite{liu2021group} for finer-grained scene encoding and adopted BERT~\cite{kenton2019bert} for query encoding. Fusion was further streamlined by directly integrating features via a text-to-3D transformer~\cite{ye2021tvcg3dqa}, enabling more direct question-to-answer mappings. Similarly, TransVQA3D~\cite{yan2023comprehensive} enhanced feature interaction by introducing SGAA for fusion, focusing on global and local semantics in scenes.

For the datasets requiring spatial grounding, FE-3DGQA~\cite{zhao2022toward} advanced the pipeline by using PointNet++~\cite{qi2017pointnet++} for spatial feature extraction and T5~\cite{raffel2020exploring} for textual encoding, complemented by an attention mechanism~\cite{zhao20213dvg, liu2021swin} to align text with dense spatial annotations.
The recently proposed SIG3D~\cite{man2024situational} focuses on context-aware tasks in embodied intelligence. It encodes scenes using voxel-based tokenization and employs anchor-based contextual estimation to determine the agent's position and orientation. 

\vspace{1mm}
{\noindent\textbf{{Multi-view and 2D-3D Methods:}}} 
A few methods also use multi-view images to enhance 3D SQA performance. For example, 3D-CLR~\cite{hong20233d} constructs compact 3D scene representations by leveraging multi-view images and optimizing 3D voxel grids. The model achieves alignment between 3D voxel features and 2D per-pixel features, grounding concepts using CLIP~\cite{radford2021learning}, which facilitates zero-shot semantic understanding. 
On the other hand, 2D-3D methods like BridgeQA~\cite{mo2024bridging} combine 2D image features from pretrained VLMs~\cite{radford2021learning, li2023blip} with 3D object-level features obtained through VoteNet~\cite{qi2019deep}. Both feature types are aligned with text features encoded by the VLM’s text encoder and fused using a vision-language transformer, enabling  free-form answers. 
\vspace{1mm}
\noindent\textbf{Advances in Text Encoders:}
The evolution of text encoders in 3D SQA reflects the increasing demands for contextual and multimodal understanding by the models. Early methods employed BiLSTM~\cite{graves2012long} and BERT~\cite{kenton2019bert} for basic semantic and syntactic feature extraction, as seen in ScanQA~\cite{azuma2022scanqa} and 3DQA-TR~\cite{ye2021tvcg3dqa}. More recent approaches, such as FE-3DGQA~\cite{zhao2022toward}, leverage  transformer-based models like T5~\cite{raffel2020exploring} for richer linguistic embeddings. Meanwhile, multimodal models like CLIP~\cite{radford2021learning} in 3D-CLR~\cite{hong20233d} and BLIP~\cite{li2023blip} in BridgeQA~\cite{mo2024bridging} have been instrumental in aligning textual and visual features. These advancements highlight a shift towards models that seamlessly integrate text with 3D spatial representations for improved performance.
\textcolor{black}{Task-specific methods are typically evaluated on the ScanQA and SQA3D datasets. Tables \ref{table4} and \ref{table5} in the appendix provide performance comparison summaries on these dataset for existing methods.}

\begin{figure*}[t]
    \centering
\includegraphics[width=0.95\linewidth]{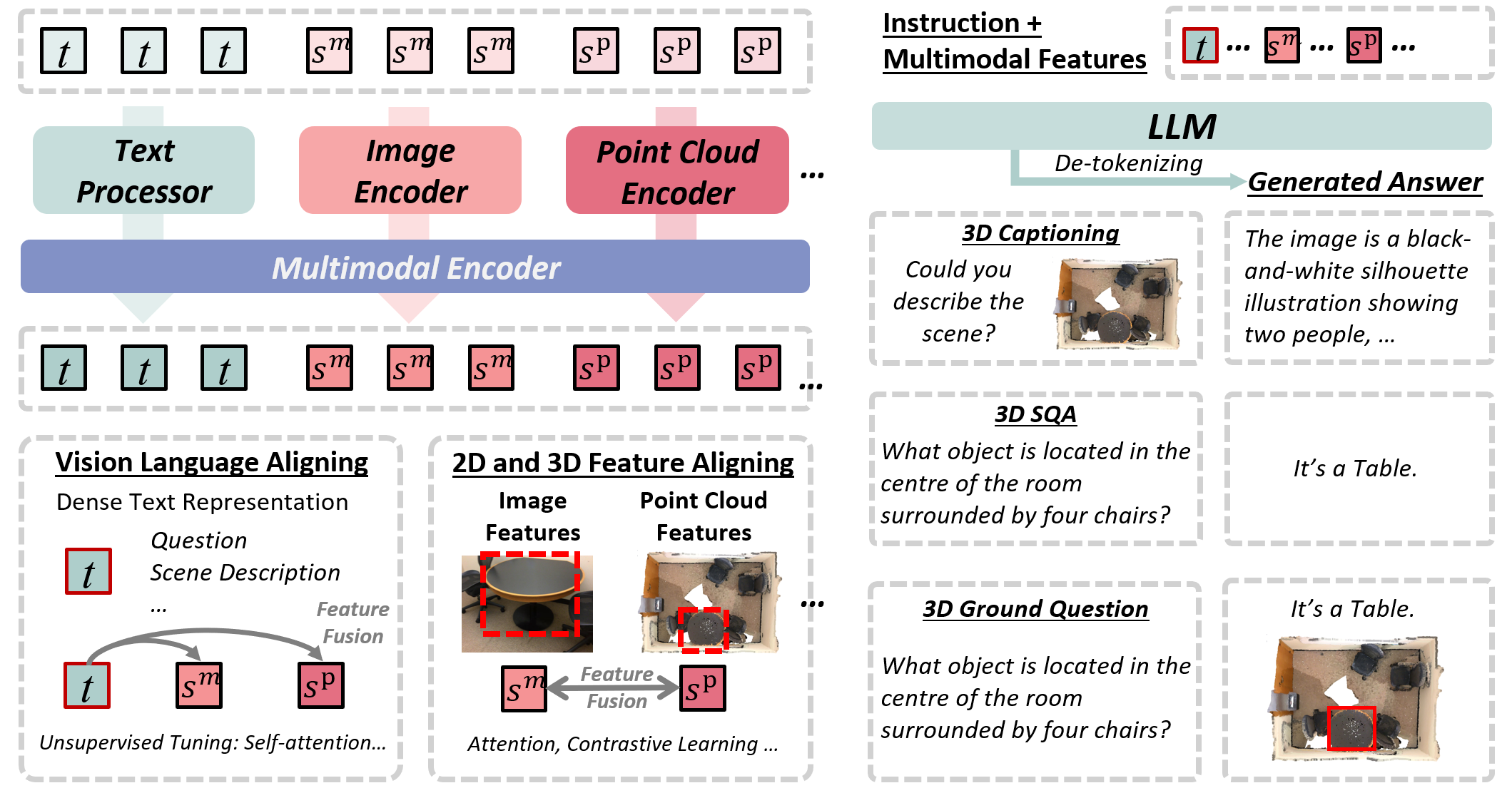}
  \caption{\textcolor{black}{Illustration of an instruction-tuned 3D SQA framework. The system encodes multimodal inputs—including text, multi-view images, and point clouds—into a shared embedding space. Through alignment modules and fusion strategies, these features are combined with task-specific instructions and forwarded to a Large Language Model (LLM). The LLM generates context-aware answers for diverse 3D tasks such as captioning, question answering, and spatial grounding, enabling generalization without task-specific heads.}}
    \label{3dvqa}
\end{figure*} 

\subsection{Pretraining-Based Methods}

Pretraining-based approaches in 3D SQA have transitioned from traditional methods that emphasize explicit alignment of spatial and textual embeddings to instruction-tuning paradigms that harness large pretrained models. These  methods strike a balance between task-specific adaptation and generalization to  address challenges of scalability. 

\vspace{1mm}
\noindent\textbf{Traditional Pretraining Methods:}
These methods focus on aligning 3D spatial features with rich 2D visual and linguistic representations. \citet{parelli2023clip} utilized a trainable 3D scene encoder based on VoteNet~\cite{qi2019deep} to extract object-level features, which are further refined using a Transformer layer to model inter-object relationships. 
Multi-CLIP~\cite{delitzas2023multi} introduces multi-view rendering and robust contrastive learning to enhance the integration of 3D spatial features with 2D representations. \citet{zhang2024vision} introduced object-level cross-contrastive and self-contrastive learning tasks during pretraining to improve cross-modal alignment. 
\citet{jia2025sceneverse} adopted a hierarchical contrastive alignment strategy, combining object-level, scene-level, and referential embeddings to enhance cross-modal and intra-modal feature integration.

 Diverging from these contrastive learning approaches, 3D-VisTA~\cite{zhu20233d} employs a unified Transformer-based framework~\cite{vaswani2017attention} to align 3D scene features with textual representations. Instead of relying on extensive annotations, it leverages self-supervised objectives to optimize multimodal alignment~\cite{he2021transrefer3d, radford2019language}.
 This shift from task-specific pretraining to self-supervised learning is a noteworthy development for   efficient and robust 3D SQA.

\vspace{1mm}
\noindent\textbf{Instruction-Tuning Methods:}
Pretrained foundation models learn general geometric and semantic representations from large-scale unsupervised data at high  computational cost.
Instruction-tuning methods exploit the generalization abilities of these models by leveraging pretrained LLMs or VLMs as frozen encoders.
These methods retain the parameters of the  encoders, making minimal modifications, typically through lightweight task-specific layers, to adapt to downstream tasks.As illustrated in Figure~\ref{3dvqa}, a typical instruction-tuned 3D SQA pipeline processes multimodal inputs (e.g., text, images, point clouds) through dedicated encoders, aligns them with task prompts, and feeds the fused representation into an LLM for answer generation. This structure supports tasks such as 3D captioning, VQA, and grounded QA in a unified manner.

LM4Vision~\cite{pang2023frozen} employs a frozen LLaMA~\cite{touvron2023llama} encoder and trains lightweight task-specific layers for alignment with the 3D QA tasks. Similarly, 3D-LLM builds upon the BLIP2~\cite{li2023blip}, adding a task-specific head while keeping the base model frozen. In contrast, LEO, M3DBench, and LAMM utilize Vicuna~\cite{chiang2023vicuna}, a derivative of LLaMA, to integrate textual and multimodal inputs. LEO incorporates object-centric  and scene-level captions for enhanced multimodal reasoning. 
LL3DA\cite{chen2024ll3da} introduces the Interactor3D module with self-attention and multimodal transformers to align scene and language features. HIS-GPT\cite{zhao2025his} enhances human-in-scene understanding through auxiliary interaction modeling and spatial trajectory encoding. View2Cap\cite{yuan2025empowering} grounds textual descriptions by explicitly predicting the observer’s position and orientation. Chat-Scene\cite{huang2024chat} reformulates multiple 3D tasks into a unified question-answering format via object identifier representations, enabling efficient instruction tuning across tasks. Scene-LLM~\cite{fu2024scene} further supports dynamic interaction by incorporating both egocentric and global scene features into a language-conditioned planning framework. \textcolor{black}{SplatTalk ~\cite{thai2025splattalk} constructs a 3D-language Gaussian field and aligns voxel representations with language through a self-supervised 3D-language Gaussian splatting training framework.}

By leveraging the extensive knowledge encoded in LLMs or VLMs, these methods bypass the need for large task-specific pretraining datasets. 
Additionally, instruction-tuning methods are also effective in zero- and few-shot scenarios. 

\begin{figure*}[t]
    \centering
\includegraphics[width=0.95\linewidth]{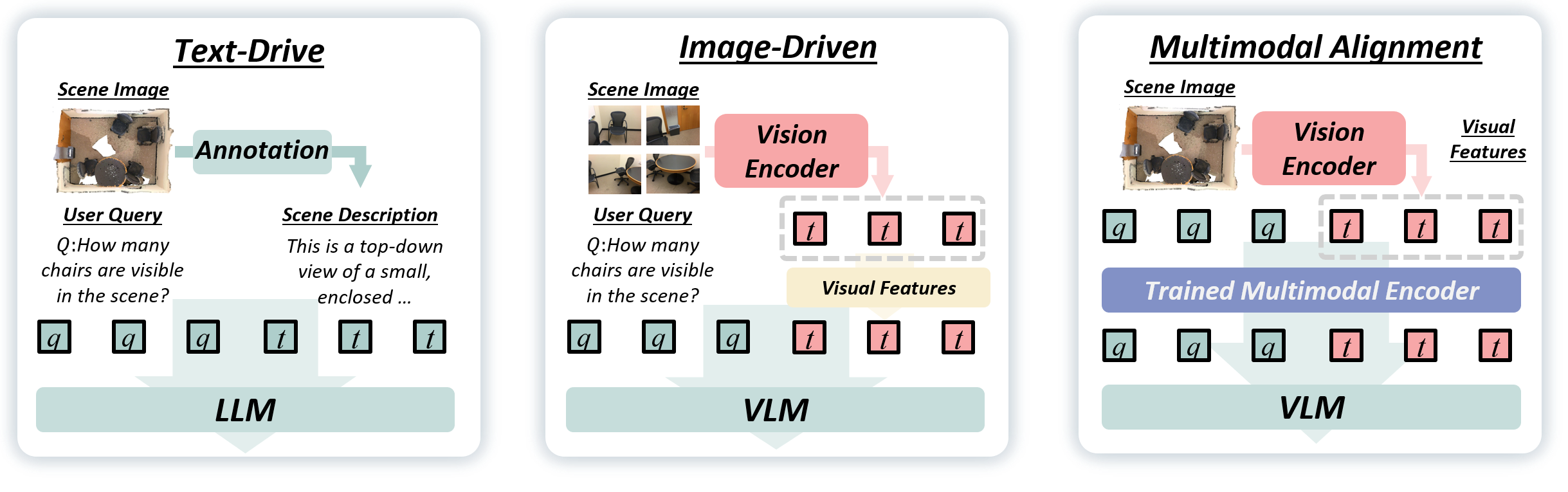}
\caption{
Illustration of three representative paradigms for zero-shot 3D SQA. 
\textbf{Left:} \textit{Text-Driven} methods (e.g., MLLM-based) use scene descriptions obtained from human annotations or external datasets, which are processed together with user queries via LLMs. 
\textbf{Middle:} \textit{Image-Driven} approaches use pretrained vision encoders to extract features from visual inputs, which are then fused with text queries by a vision-language model (VLM). 
\textbf{Right:} \textit{Multimodal Alignment} strategies employ pretrained multimodal encoders trained on large-scale vision-language datasets and apply them directly to 3D SQA without task-specific fine-tuning, relying on the generalization of aligned representations.
}
    \label{zero}

\end{figure*} 

\subsection{Zero-Shot Learning Methods}

Zero-shot has emerged as a promising learning paradigm for 3D SQA, enabling models to infer answers to unseen tasks without task-specific fine-tuning. Current zero-shot 3D SQA methods can be broadly categorized into: text-driven, image-driven, and multimodal alignment approaches, as illustrated in Figure~\ref{zero}.

\vspace{1mm}
\noindent\textbf{Text-Driven Approaches:} These methods convert 3D scene information into textual descriptions, which are then used with a question in pretrained LLMs or VLMs for zero-shot inference. An example is SQA3D~\cite{ma2022sqa3d}, which uses Scan2Cap~\cite{chen2021scan2cap} to generate scene descriptions and inputs them into GPT-3~\cite{brown2020language} for answering questions. However, this approach overlooks the spatial structure of point clouds and images, limiting its ability to fully leverage 3D information. Similarly, LAMM~\cite{yin2024lamm} extracts features from point clouds and text, but uses 3D data in a limited manner.  

\vspace{1mm}
\noindent\textbf{Image-Driven Approaches:} These methods use VLMs to incorporate visual features like images or multi-view data along with text. For instance, MSQA~\cite{linghu2024multi} uses GPT-4o~\cite{achiam2023gpt} with VLMs.  
\citet{singh2024evaluating} tested unfinetuned GPT-4V~\cite{yang2023dawn} on datasets like 3D-VQA and ScanQA~\cite{azuma2022scanqa}, showing competitive performance in certain tasks. These methods are flexible and resource-efficient, but they  still rely on text to represent spatial and object relationships, which is a potential limitation. 

\vspace{1mm}
\noindent\textbf{Multimodal Alignment Approaches:} Techniques such as LEO~\cite{huang2023embodied} and Spartun3D-LLM~\cite{zhang2024spartun3d}, explicitly align visual and textual information during pretraining. LEO  improves zero-shot performance by aligning object- and scene-level features, while Spartun3D-LLM employs an  explicit module for aligning point clouds and text.
These methods require relatively more training resources due to additional computations. Nevertheless, they offer an attractive trade-off between performance and efficiency.  

Overall, in contemporary Zero-shot 3D SQA, Text-driven approaches are cost-effective and flexible but suffer from limited utilization of 3D data. Image-driven methods, which directly leverage VLMs for inference, also face limitations due to insufficient exploitation of 3D information. Multimodal alignment methods, while offering superior performance, have higher resource requirements. 

\setlength{\tabcolsep}{4pt} 
\renewcommand{\arraystretch}{1.2} 
\begin{table*}[htp]
\centering
\caption{Performance comparison of existing models on ScanQA datasets. \textbf{EM@1} and \textbf{EM@10} refer to exact match accuracy for top-1 and top-10 answers, respectively. \textbf{B-1} to \textbf{B-4} represent BLEU-1 to BLEU-4 scores. \textbf{R}, \textbf{M}, and \textbf{C} stand for ROUGE, METEOR, and CIDEr metrics, respectively. 
}
\small
\begin{tabular}{p{2.4cm}|p{1.4cm}p{1.1cm}p{1.1cm}p{1.1cm}p{1.1cm}p{1.1cm}p{1.1cm}p{1.1cm}p{1.1cm}p{1.1cm}}
\hline
\textbf{Model} & \textbf{Type} & \textbf{EM@1} & \textbf{EM@10} & \textbf{B-1} & \textbf{B-2} & \textbf{B-3} & \textbf{B-4} & \textbf{R} & \textbf{M} & \textbf{C} \\ \hline
ScanQA~\cite{azuma2022scanqa} & T-S & 21.05 & 51.23 & 30.24 & 20.40 & 15.11 & 10.08 & 33.30 & 13.14 & 64.90 \\
FE-3DGQA~\cite{zhao2022toward} & T-S & 22.26 & 54.51 & -     & -     & -     & -     & -     & -     & -     \\

SIG3D~\cite{man2024situational} & T-S & -     & -     & 39.50 & -     & -     & 12.40 & 35.90 & 13.40 & 68.80 \\
\hline
ESZG~\cite{singh2024evaluating} & Z-S & 18.01 & 18.01 & 30.24 & 20.40 & 15.11 & 10.08 & 33.33 & 13.14 & 64.86 \\ \hline
3DVLP~\cite{zhang2024vision} & P-B & 24.03 & 57.91 & -     & -     & -     & -     & -     & -     & -     \\
CLIP-Guided~\cite{parelli2023clip} & P-B & 23.92 & -     & 32.82 & -     & -     & 14.64 & 35.15 & 13.94 & 69.53 \\
Multi-CLIP~\cite{delitzas2023multi} & P-B & 24.02 & -     & 32.63 & -     & -     & 12.65 & 35.46 & 13.97 & 68.70 \\
3D-VisTA~\cite{zhu20233d} &  P-B & 27.00 & 57.90 & -     & -     & -     & 16.00 & 38.60 & 15.20 & 76.60 \\
LAMM~\cite{yin2024lamm} & P-B(w I-T) & -     & -     & -     & -     & -     & -     & -     & -     & -     \\
3D-LLM~\cite{hong20233d3dllm} & P-B(w I-T) & 21.20 & -     & 39.30 & 25.20 & 18.40 & 12.00 & 37.85 & 15.10 & 74.50 \\
\textcolor{black}{SplatTalk} ~\cite{thai2025splattalk} & P-B(w I-T)& 22.40     & -     & - & -     & -     & - & - & - & - \\ 
SceneVerse~\cite{jia2025sceneverse} & P-B(w I-T) & 22.70 & -     & -     & -     & -     & -     & -     & -     & -     \\

\textcolor{black}{LL3DA} ~\cite{chen2024ll3da} & P-B(w I-T) & -     & -     & - & -     & -     & 13.53 &  37.31 & 15.88 & 76.79 \\

\textcolor{black}{Scene-LLM}~\cite{fu2024scene}  & P-B(w I-T) & 27.20     & -     & 43.60 & 26.80     & 19.10     & 12.00 &  40.00 &16.60 & 80.00 \\
\textcolor{black}{Chat-Scene}~\cite{huang2024chat}  & P-B(w I-T) & -     & -     & - & -     & -     & 14.30 &  - & - & 87.70 \\\hline

Human & - & 51.60 & -     & -     & -     & -     & -     & -    & -     &-     \\ \hline
\end{tabular}
\label{table4}
\end{table*}

\begin{table*}[h]
\centering
\caption{Performance comparison of existing models on SQA3D datasets. The question types include "What," "Is," "How," "Can," "Which," and "Others," with the "Avg" column representing the average performance across all types. The metric used is accuracy. 
}
\small
\setlength{\tabcolsep}{4pt} 
\begin{tabular}{l|cccccccc}
\hline
\textbf{Model} & \textbf{Type} & \textbf{What} & \textbf{Is} & \textbf{How} & \textbf{Can} & \textbf{Which} & \textbf{Others} & \textbf{Avg} \\ \hline
ScanQA~\cite{azuma2022scanqa}                 & T-S & 28.60 & 65.00 & 47.30 & 66.30 & 43.90 & 42.90 & 45.30 \\
SQA3D~\cite{ma2022sqa3d}                      & T-S & 33.48 & 66.10 & 42.37 & 69.53 & 43.02 & 46.40 & 47.02 \\
SIG3D~\cite{man2024situational}              & T-S  & 35.60 & 67.20 & 48.50 & 71.40 & 49.10 & 45.80 & 52.60 \\ \hline
SQA3D (Z-S)~\cite{ma2022sqa3d}              & Z-S & 39.67 & 45.99 & 40.47 & 45.56 & 36.08 & 38.42 & 41.00 \\
\hline

Multi-CLIP~\cite{delitzas2023multi}           & P-B & -     & -     & -     & -     & -     & -     & 48.00 \\
3D-VisTA~\cite{zhu20233d}                     & P-B & 34.80 & 63.30 & 45.40 & 69.80 & 47.20 & 48.10 & 48.50 \\
3D-LLM~\cite{hong20233d3dllm}                 & P-B(w I-T) & 35.00 & 66.00 & 47.00 & 69.00 & 48.00 & 46.00 & 48.10 \\

LM4Vision~\cite{pang2023frozen}               & P-B(w I-T) & 34.27 & 67.05 & 48.17 & 68.34 & 43.87 & 45.64 & 48.10 \\
SceneVerse~\cite{jia2025sceneverse}           & P-B(w I-T) & -     & -     & -     & -     & -     & -     & 49.90 \\
LEO~\cite{huang2023embodied}                  & P-B(w I-T) & 46.80 & 64.10 & 47.00 & 60.80 & 44.20 & 54.30 & 52.90 \\
\textcolor{black}{Scene-LLM}~\cite{fu2024scene} & P-B(w I-T) & 40.90  & 69.10  & 45.00  & 70.80  & 47.20  & 52.30  & 54.20  \\
Spartun3D-LLM~\cite{zhang2024spartun3d}       & P-B(w I-T) & 49.40 & 67.30 & 47.10 & 63.40 & 45.40 & 56.60 & 54.90 \\
\hline
Human                                         & -  & 88.53 & 93.84 & 88.44 & 95.27 & 87.22 & 88.57 & 90.06 \\ \hline
\end{tabular}
\label{table5}
\end{table*}

\subsection{State-of-the-arts}
\textcolor{black}{Despite the recent emergence of numerous 3D SQA benchmarks, there remains a lack of general-purpose and widely adopted datasets that enable a fair and comprehensive comparison across all method categories. To address this gap, we focus our evaluation on ScanQA~\cite{azuma2022scanqa}  and SQA3D~\cite{ma2022sqa3d}, two of the most widely used and representative datasets in current 3D SQA research. These benchmarks provide a common ground for assessing the strengths and weaknesses of different approaches under consistent settings. Specifically, Table~\ref{table4} summarizes the performance of various methods on the ScanQA dataset, while Table~\ref{table5} presents a detailed breakdown by question type on the SQA3D dataset. Together, these results offer valuable insights into the comparative effectiveness and applicability of existing methods.}

The ScanQA results (Table~\ref{table4}) show that instruction-tuned pretraining-based models deliver the strongest performance. Scene-LLM~\cite{fu2024scene} achieves the highest EM@1 (27.20) and a CIDEr score of 80.00, while Chat-Scene~\cite{huang2024chat} further improves CIDEr to 87.70, indicating strong alignment between language generation and 3D spatial context. Pretraining-based models without instruction tuning, such as 3D-VisTA~\cite{zhu20233d}, also perform well (EM@1 = 27.00), benefiting from large-scale scene-language representation learning. In contrast, task-specific methods (e.g., ScanQA~\cite{azuma2022scanqa}, SIG3D~\cite{man2024situational} ) demonstrate decent accuracy (up to 22.4 EM@1) but generally lag behind in generation quality.
On the SQA3D benchmark (Table~\ref{table5}), similar trends are observed. Instruction-tuned models, such as Spartun3D-LLM~\cite{zhang2024spartun3d}   and Scene-LLM~\cite{fu2024scene}, outperform other approaches with average accuracy above 54, showing consistent advantages across diverse question types. Pretraining-based methods (e.g., 3D-LLM~\cite{hong20233d3dllm} , LM4Vision~\cite{pang2023frozen}  ) follow closely, achieving stable performance without relying on dataset-specific designs. Task-specific models, while effective in some categories (e.g., "Is" or "Can" questions), remain less flexible and show limited gains overall.

In both benchmarks, zero-shot methods (e.g., ESZG~\cite{singh2024evaluating}, SQA3D(Z-S)~\cite{ma2022sqa3d}) perform the worst, with significantly lower scores across all metrics. Although these models offer scalability and generalization potential, they currently struggle to capture fine-grained spatial understanding, highlighting the need for future research in effective zero-shot adaptation for 3D QA tasks.

In summary, instruction-tuned multimodal models represent the most effective solution for 3D Scene Question Answering, combining strong scene-language alignment with flexible task transferability. Pretraining-based methods without instruction tuning offer competitive baselines, while task-specific designs provide valuable insights but lack generalizability. Zero-shot approaches remain promising in principle, yet still fall short in addressing the full complexity of grounded 3D reasoning. Importantly, despite recent progress, all existing methods still fall short of human-level performance—especially on open-ended and multi-step reasoning tasks.  This indicates that while significant gaps remain, the field is progressing rapidly and holds considerable room for future advancement, particularly in building more generalizable, spatially aware, and instruction-following agents.

\section{Challenges and Future Directions\label{sec6}}
While 3D SQA has seen notable advancements, several critical challenges remain, limiting its potential for real-world applications. We outline key challenges and propose directions for future research.

\vspace{1mm}
\noindent\textbf{Dataset Quality and Standardization.} The rapid development of 3D SQA datasets in recent years has led to a fragmented landscape, with datasets varying widely in scope and modality. Integrating these datasets into unified benchmarks can offer the much needed standardised evaluation to catapult research in this direction. 
Additionally, while LLMs facilitate scalable dataset generation, they often introduce hallucinated information and contextual misalignments. Future research should focus on robust validation frameworks, leveraging human-in-the-loop systems or LLMs as validators. 

\vspace{1mm}
\noindent\textbf{Enhancing 3D Awareness in Zero-Shot.} Current zero-shot models heavily rely on textual proxies, with limited utilization of 3D spatial and geometric features. Although multi-view approaches mitigate this issue to some extent, the lack of explicit 3D representation hampers their effectiveness for spatially complex tasks. Instruction-tuning methods face similar limitations. Future work needs to explore architectures that deeply integrate 3D features with linguistic and visual modalities to enhance generalization across diverse tasks. Additionally, an apparent  direction for future research is to  explore the balance between multimodal alignment and pretrained models in zero-shot 3D SQA to enhance both efficiency and performance. 

\vspace{1mm}
\noindent\textbf{Unified Evaluation.} Absence of standardized and 3D SQA objective-specific evaluation metrics currently complicates meaningful  evaluation and comparisons across datasets and models. Developing unified frameworks that incorporate multimodal metrics for spatial reasoning, contextual accuracy, and task-specific performance are currently required to enable accurate benchmarking and drive methodological innovation in 3D SQA.

\vspace{1mm}
\noindent\textbf{Dynamic and Open-World Scenarios.} Most existing methods and datasets focus on static, predefined environments, limiting applicability to real-world tasks. Future efforts need to emphasize more on dynamic, open-world settings, enabling models to handle real-time scene changes and novel queries. Incorporating embodied interactions, such as navigation and multi-step reasoning, will further align 3D SQA systems with real-world requirements.

\vspace{1mm}
\noindent\textbf{Interpretable and Explainable 3D SQA Models.}
Current 3D SQA models often act as "black boxes", limiting their adoption in trust-critical domains like healthcare. Developing interpretable models that visualize 3D features, highlight relevant regions, or provide natural language explanations can enhance user trust and broaden their applicability.

\vspace{1mm}
\noindent\textbf{Multimodal Interaction and Collaboration.}
3D SQA systems are evolving toward more natural and interactive interfaces. Future research can explore integrating linguistic, gestural, and visual inputs to enable intuitive interaction with 3D scenes. Additionally, collaborative scenarios, such as architectural design or educational training, where multiple users interact with the system in real-time, offer a promising direction. Such systems could enhance communication and joint problem-solving, unlocking broader applications for 3D SQA.

\vspace{1mm}
\noindent\textbf{Incorporating Temporal Dynamics.}
Most 3D SQA models currently ignore temporal dynamics of the scenes, whereas most of the real-world applications, such as traffic monitoring, robotic navigation,  involve dynamic environments. Future research should aim to incorporate temporal dynamics into 3D SQA, allowing models to reason about scene changes over time. Leveraging temporal information, such as object movements, would enable these systems to better handle tasks requiring long-term temporal reasoning.

\vspace{1mm}
\noindent\textbf{Model Efficiency and Deployment.}
\textcolor{black}{Deploying 3D SQA systems on resource-constrained devices, such as mobile robots and edge AI agents, remains challenging due to high computational and memory demands. Future work should focus on lightweight architectures and optimization techniques, including pruning, quantization, and knowledge distillation, to enable efficient and real-time inference. Energy-efficient algorithms and scalable designs tailored for embedded systems will further enhance the practicality of 3D SQA in real-world applications.}

\vspace{1mm}
\noindent\textbf{Practical Deployment and Application Challenges.}
\textcolor{black}{3D SQA has promising applications in household robotics, AR/VR training, and warehouse automation. A service robot may answer queries like “What is on the kitchen counter?” to assist with navigation or object retrieval. In AR/VR, users can explore virtual scenes by asking spatial questions such as “How many chairs are in this room?”. In warehouses, robots with persistent 3D memory can quickly locate objects or respond to “Where is the red toolbox now?” and “How many packages remain on Shelf A?”, enabling efficient inventory management.
However, currently, real-world deployment faces notable challenges. The domain gap between curated datasets and complex environments affects model generalization. Systems must operate in real time under limited resources, handle ambiguous or incomplete queries, and maintain consistent multi-modal perception. In practical settings, audio-based interaction (e.g., voice commands) and multi-turn dialogue are also essential, requiring models to understand spoken language, retain conversational context, and respond coherently over time. Bridging these gaps is critical for developing robust and scalable 3D SQA systems ready for real-world use.}

\vspace{1mm}
\noindent\textbf{Interdisciplinary Collaboration and Integration Opportunities.}
\textcolor{black}{Interdisciplinary knowledge plays a crucial role in advancing 3D SQA. Cognitive science provides insights into human spatial reasoning, such as gaze-based attention, mental simulation, and context-dependent grounding, which can guide more human-aligned perception and question understanding. Human-computer interaction (HCI) contributes to the design of intuitive interfaces, natural interaction protocols, and evaluation criteria focused on usability and communicative effectiveness.
Importantly, 3D SQA in real-world settings often involves not only language understanding but also interpretation of human actions—such as pointing, approaching, or manipulating objects—which are crucial for resolving spatial references and supporting task-oriented interaction. Modeling such embodied queries currently remains an open challenge and a key direction for future research.
Integrating these interdisciplinary perspectives is non-trivial due to the difficulty of formalizing qualitative knowledge, aligning cross-domain evaluation standards, and building shared benchmarks. Bridging these gaps will foster more interactive, robust, and cognitively grounded 3D SQA systems.}


By addressing these challenges, 3D SQA can advance toward robust, scalable, and versatile systems, accelerating real-world deployment and driving progress in embodied intelligence and multimodal understanding.

\section{\textcolor{black}{Conclusion}\label{sec7}} 
This survey presents a comprehensive overview of 3D Scene Question Answering (3D SQA), a rapidly evolving field at the intersection of 3D computer vision and natural language processing. 3D SQA plays a pivotal role in advancing embodied intelligence by enabling spatial understanding and multimodal reasoning.
We reviewed the evolution of datasets—from manual curation to LLM-assisted generation—and the progression of methods from task-specific pipelines to zero-shot paradigms. Through systematic categorization of datasets and methodologies, we identified their respective strengths and limitations, and analyzed the need for more unified evaluation protocols and data construction standards.
To address persistent challenges such as dataset quality, multimodal alignment, and evaluation consistency, we outlined promising research directions and emerging trends. We hope this work provides a foundation for further exploration, supporting the development of robust and scalable systems capable of handling complex real-world 3D tasks.



\printcredits

\bibliographystyle{cas-model2-names}

\bibliography{reference}

\end{document}